\documentclass{article}

\usepackage{microtype}
\usepackage{graphicx}
\usepackage{subfigure}
\usepackage{booktabs} 
\usepackage{multirow}
\usepackage{array}    
\usepackage{listings}
\usepackage{enumitem}

\usepackage{hyperref}
\usepackage[table]{xcolor} 
\usepackage{lmodern}
\usepackage{algorithm}
\usepackage{algpseudocode}
\usepackage{amsmath,amssymb} 

\usepackage[accepted]{icml2024}

\usepackage{amsmath}
\usepackage{amssymb}
\usepackage{mathtools}
\usepackage{amsthm}

\usepackage[capitalize,noabbrev]{cleveref}

\theoremstyle{plain}

\theoremstyle{definition}

\theoremstyle{remark}

\usepackage[textsize=tiny]{todonotes}

\icmltitlerunning{Self-Regulation and Requesting Interventions}

\begin{document}

\twocolumn[
\icmltitle{Self-Regulation and Requesting Interventions}



\icmlsetsymbol{equal}{*}

\begin{icmlauthorlist}
\icmlauthor{So Yeon Min}{yyy}
\icmlauthor{Yue Wu}{yyy}
\icmlauthor{Jimin Sun}{yyy}
\icmlauthor{Max Kaufmann}{zzz}
\icmlauthor{Fahim Tajwar}{yyy}
\icmlauthor{Yonatan Bisk}{yyy}
\icmlauthor{Ruslan Salakhutdinov}{yyy}
\end{icmlauthorlist}

\icmlaffiliation{yyy}{Carnegie Mellon University, Pittsburgh, United States}
\icmlaffiliation{zzz}{University of Toronto, Toronto, Canada}

\icmlcorrespondingauthor{So Yeon Min}{soyeonm@andrew.cmu.edu}

\icmlkeywords{Machine Learning, ICML}

\vskip 0.3in
]



\printAffiliationsAndNotice{} 

\begin{abstract}
Human intelligence involves metacognitive abilities like self-regulation, recognizing limitations, and seeking assistance only when needed. While LLM Agents excel in many domains, they often lack this awareness. Overconfident agents risk catastrophic failures, while those that seek help excessively hinder efficiency. A key challenge is enabling agents with a limited intervention budget $C$ is to decide when to request assistance. 
In this paper, we propose an offline framework that trains a ``helper'' policy to request interventions, such as more powerful models or test-time compute, by combining LLM-based process reward models (PRMs) with tabular reinforcement learning. Using state transitions collected offline, we score optimal intervention timing with PRMs and train the helper model on these labeled trajectories. This \textit{offline} approach significantly reduces costly intervention calls during training. Furthermore, the integration of PRMs with tabular RL enhances robustness to off-policy data while avoiding the inefficiencies of deep RL. We empirically find that our method delivers optimal helper behavior.
\end{abstract}

\section{Introduction}
\label{submission}

\begin{figure}[!t]
    \centering
    \includegraphics[width=1.0\columnwidth]{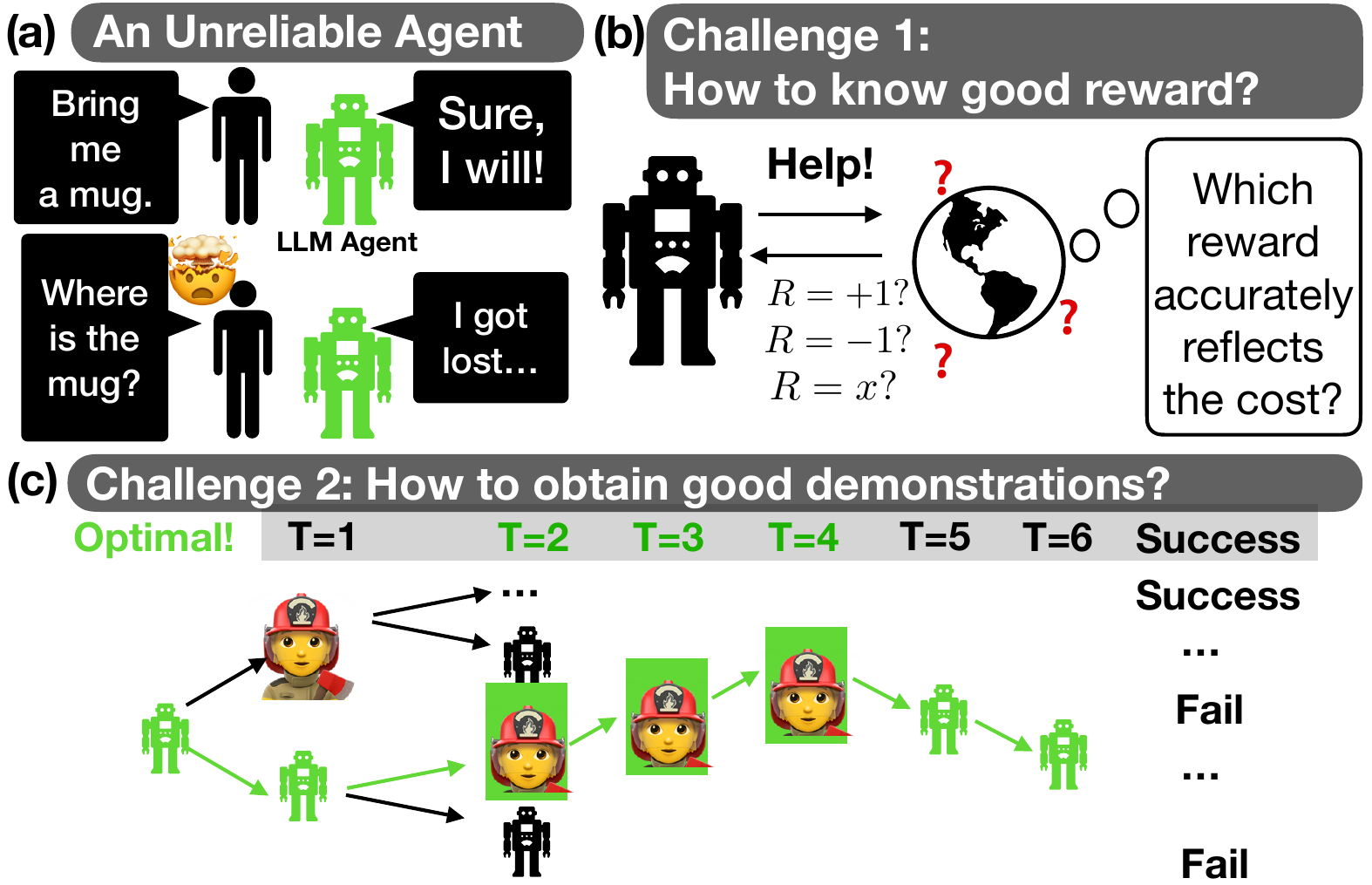}
    \vspace{-2.5em}
    \caption{\textbf{Unreliable agents and training challenges.}
    \textbf{(a)} An unreliable agent neither completes the assigned task nor communicates its inability, causing surprise and catastrophe.  \textbf{(b)} When there is a budget $C$ on interventions requested during inference, a key challenge is determining a reward function that guides the agent to request help appropriately. \textbf{(c)} For both behavior cloning/reinforcement learning, obtaining an optimal demonstration is complicated by the exponential space of possible trajectories, difficult even with human effort.}
    \vspace{-1em}
    \label{fig:fig1}
\end{figure}

Human intelligence is distinguished by metacognitive abilities, particularly self-regulation --- the capacity to monitor limitations --- and requesting targeted assistance only when needed (Fig~\ref{fig:fig1}). By recognizing and communicating uncertainties, individuals can delegate tasks or seek help before failure becomes inevitable. This approach prevents costly mistakes and fosters trust, as admitting uncertainty and asking for help at the right time is reassuring.

Despite advancements in Large Language Models (LLMs), current AI agents often lack metacognitive awareness. Overconfident agents risk catastrophic errors, while those seeking help excessively are inefficient. Ideally, an AI agent should gauge uncertainty and selectively request assistance, ensuring reliability and efficient use of human effort. While existing AI safety research addresses unintended and malicious behaviors \citep{gabriel2020artificial,christiano2017deep,bai2022constitutional}, reliability with true agency also requires the ability to recognize and communicate limitations.

A key challenge in training an intervention-requesting agent within a limited intervention budget $C$ is deciding when to request help. This involves balancing \textit{reward design} and \textit{policy optimization} (Fig.~\ref{fig:fig1}(b)): overly incentivizing help requests exhausts the budget prematurely, while under-incentivizing them leads to avoiding help altogether. Designing effective reward functions is non-trivial and can require the costly iterations of training, evaluation, and adjustment. Similarly, generating annotated trajectories for supervised fine-tuning under budget constraints is resource-intensive (Fig.~\ref{fig:fig1}(c)), as the trajectory space is exponentially large, and even human annotators can struggle to identify optimal intervention timing for every different budget constraint. 

Additionally, collecting data for policy training can be highly expensive, especially when real-world interventions rely on human effort or costly external services. This raises important questions: Can we perform intervention-based data collection once and reuse it across multiple budget constraints? Furthermore, can we also control the number of interventions during training? Even if we train a policy to adhere to the budget $C$ is during inference, managing intervention usage during training is equally critical. 

To address the challenges of efficient reward-policy search and minimizing intervention costs, we propose an \textit{offline} and \textit{hybrid} framework that collects intervention data in a single pass and combines deep scoring functions with classical tabular reinforcement learning (RL). This three-step method (Fig.~\ref{fig:method}) integrates LLM-based process reward models (PRMs) with tabular RL to enable efficient trajectory generation and policy optimization under both \textit{inference} and \textit{training} budget constraints:

\begin{enumerate}[leftmargin=1em, itemsep=5pt, parsep=0pt, topsep=2pt, partopsep=0pt]
    \vspace{-0.5em}
    \item \textbf{Transition Model Collection and PRM Training}: We collect state transitions with randomly triggered interventions, approximating environment dynamics. PRMs for the base actor and intervention are learned.\vspace{-0.5em}
    \item \textbf{Iterative Reward and Usage/Policy Search}: We apply tabular dynamic programming (DP) with PRMs and the transition model to compute optimal trajectories that respect budget constraints. A key output of DP is the expected number of help requests from the initial state. We iterate until reaching the target budget, \textit{without} having to retrain the policy for each reward configuration.\vspace{-0.5em}
    \item \textbf{Policy Training}: The annotated trajectories are used to fine-tune the helper policy, enabling it to make effective intervention decisions within budget constraints.\vspace{-0.7em}
\end{enumerate}

This \textit{offline} approach greatly reduces costly intervention calls during training. Its \textit{hybrid} design leverages PRMs alongside tabular RL, enhancing robustness to off-policy data and avoiding the inefficiencies of deep RL. Empirical results on Situated Instruction Following tasks show that our method—using powerful models and test-time compute (e.g., MCTS) as interventions—achieves performance comparable to a system that employs interventions at every step (e.g., eight per task), yet requires far fewer interventions (e.g., only one per task). By training LLM agents to self-regulate and request assistance judiciously, we take a step toward reliable deployment of LLM-based systems.

\vspace{-0.5em}
\section{Related Work}

\textbf{LLM agents }  Recent breakthroughs in LLM agents \citep{yao2023react,yang2024sweagent,shinn2024reflexion}  have allowed the creation of AI systems which can complete a range of real-world tasks in an open-ended environment \citep{nakano2021webgpt, schick2023toolformer, openhands}.
Most previous work on training such LLM agents focuses on SFT for tool-use \citep{schick2023toolformer}, prompting closed-source LLMs \citep{yang2024sweagent,wang2024executable,openhands}, or applying RL in domains with a clear objective such as code generation or math \citep{dubey2024llama,chen2024self}. We instead focus on applying RL techniques to an environment with ambiguous instructions \citep{min2025situated}. Although previous works attempt to train LLM agents in such environments \citep{zhai2024finetuning,gehring2024rlef}, they do not address how to enable them to request interventions, which is the central focus of our work.

\textbf{Safe and Trustworthy AI } Prior work on AI safety often focuses on \emph{value alignment}—ensuring AI systems follow human values \citep{gabriel2020artificial,christiano2017deep,bai2022constitutional}—and \emph{AI security}—ensuring robustness to adversarial attacks \citep{hendrycks2021unsolved,brundage2018malicious}. However, these alone may not guarantee safety in high-stakes contexts, where an agent’s limited \emph{capabilities} can lead to harmful failures (e.g., prescribing the wrong medicine \citep{ruan2024identifyingsandbox}). We therefore situate our work under the more expansive concpet of \emph{Trustworthy AI} \citep{diaz2023connectingtrustworthy}, which includes the requirement that agents pursue tasks robustly without unintended failures. 

\begin{figure*}[!t]
    \centering
    \includegraphics[width=1.0\textwidth]{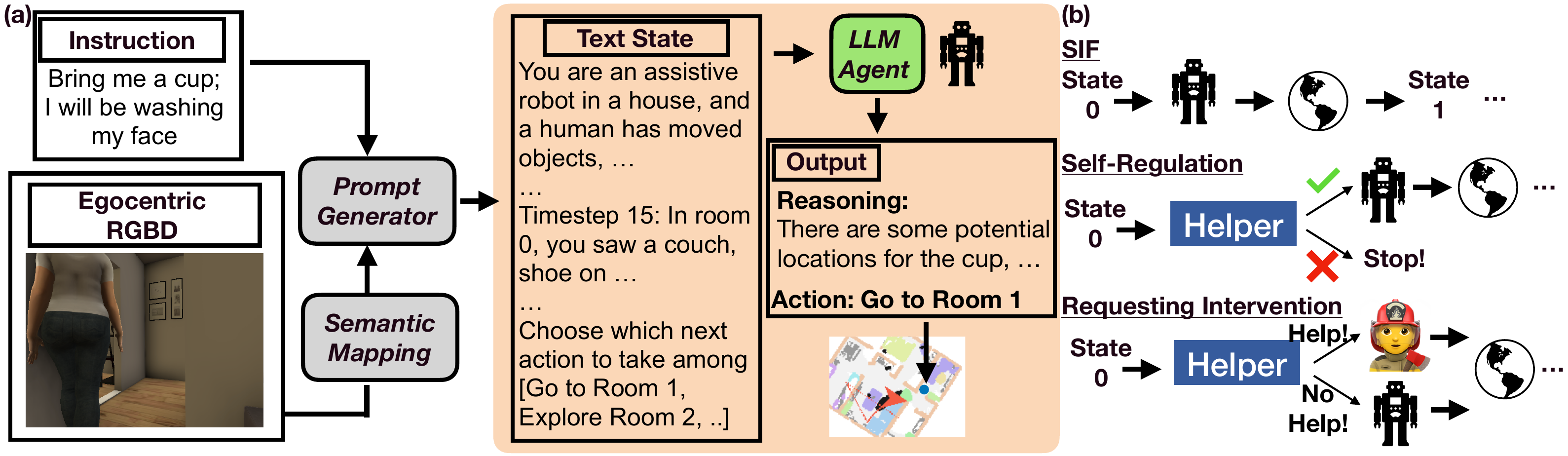}
    \vspace{-2em}
    \caption{(a) A SIF task requires the agent to locate objects, interact with humans, and perform household tasks in a sequence of discrete actions. Assuming perfect visual perception, the relevant segment is highlighted in orange; states are represented in text. (b) A brief overview of Self-Regulation and Requesting Intervention, in comparison to the base agent.}
    \label{fig:sif}
    \vspace{-1em}
\end{figure*}

\textbf{Self-improvement for LLMs }Previous work in self-improvement has explored the potential of enhancing LLM responses. 
Environmental feedback~\cite{gou2024critic,qiao-etal-2024-making,liu2024agentbench,chen2024teaching,anonymous2025livecodebench} and model-generated critiques~\cite{madaan2023selfrefine,wang2023shepherdcriticlanguagemodel,welleck2023generating,lin2024generating,qu2024recursive} have enabled models to perform better in subsequent iterations. Reward models combined with search algorithms further guide decoding toward better answers~\cite{nakano2021webgpt,uesato2022solvingmathwordproblems,zelikman2022star,xie2023selfevaluation,lightman2024lets}. However, most such methods assume the model can inherently solve the task, with the challenge lying in eliciting that capability. When a task exceeds the model’s ability, intervention of more capable models/augmented compute is needed.

\textbf{Confidence Estimation \& Meta-cognition.} Meta-cognitive agents that recognize their own limitations can guide human trust and seek external knowledge to improve accuracy~\cite{mallen-etal-2023-trust,asai2024selfrag}. Previous work estimates confidence via semantic entropy~\cite{kuhn2023semantic}, logit values~\cite{jiang-etal-2021-know,Kadavath2022LanguageM}, direct questioning~\cite{zhou-etal-2023-navigating,lin2024generating,xiong2024can}, or conformal prediction~\cite{ren2023robots}. Although these methods can help decide when to intervene, their estimates are calculated from logits, and may be biased by training data and fail out-of-distribution~\cite{Xiao2022UncertaintyQW,zhou-etal-2023-navigating}. Another approach is learning an RL policy that treats assistance seeking as an action~\cite{Chi2019JustAA,Nguyen2021LearningWA,liu2022asking,singh2022askhelp,xie2022when,hu2024uncertainty}. In contrast, we use a process reward model with tabular RL to flexibly adapt to varying budget constraints without additional training.

\section{Task and Setup}

\textbf{Task } 
We use the Situated Instruction Following (SIF) task~\cite{min2025situated}, which requires finding objects, interacting with humans, and performing household tasks in highly uncertain and dynamic environments (Fig.~\ref{fig:sif}). 
To the best of our knowledge, SIF is among the most suitable benchmarks for evaluating how well LLM-driven agents handle nuanced and uncertain instructions. The environment and the instructions become uncertain because the speaker’s intent is not always fully specified, and the human may dynamically alter the scene (e.g., placing an object in a different room or moving to a new location). Even advanced models like GPT-4o struggle with these tasks due to such inherent ambiguities~\cite{min2025situated}.

Our setup provides perfect visual perception, allowing the agent to receive textual environment descriptions at each step. Its action space includes discrete commands such as \emph{Go to Room X} and \emph{Explore Room X}, forming a multi-turn, text-based setting suitable for LLMs. Even with perfect perception, agents must calibrate their uncertainty to interpret ambiguous instructions: humans may move objects around or relocate themselves, forcing the agent to determine whether it should gather more evidence, search additional rooms, or attempt to clarify instructions.

Following SIF, we use two task types. \emph{PnP} is a straightforward pick-and-place scenario. \emph{S\_obj} is more challenging: it involves ambiguous human hints, partial observability, and the possibility of objects being moved. We use 1,000 training tasks and 40 validation/test tasks per split.

\textbf{Reliable Behaviors } We focus on two key strategies that can achieve reliable agent behaviors: self-regulation and requesting interventions (Fig.~\ref{fig:sif}(b)). \textbf{Self-regulation} involves the agent autonomously deciding to stop task execution when it cannot successfully complete the task. This prevents the agent from continuing in scenarios where failure is likely, thereby conserving resources and maintaining reliability. In contrast, \textbf{requesting interventions} refers to the agent making state-wise requests for assistance during specific states within the task, receiving help at critical points.

\textbf{Base Actors and Interventions }
We employ two base actors: \textbf{GPT-4o-mini} \citep{achiam2023gpt} and \textbf{LLaMA 3B} \citep{dubey2024llama}. LLaMA models are trained on GPT-4o and \(\text{GPT-4o-mini}\) trajectories for the \(\textit{S\_obj}\) and \(\textit{PnP}\) tasks, respectively. 
For \textbf{requesting interventions}, we employ three kinds of interventions:

\vspace{-0.2\baselineskip}
\textit{Depth-1 MCTS:} A simple Monte Carlo Tree Search approach that, guided by a process reward model, generates up to five candidate actions and selects the best. Details on MCTS implementation are in Appendix~\ref{app:mcts}.

\vspace{-0.2\baselineskip}
\textit{More Powerful Models:} For a \(\text{GPT-4o-mini}\) base actor, we invoke \(\text{GPT-4o}\). For a \(\text{LLaMA 3B}\) base actor, we use a \(\text{LLaMA 3B}\) with a better performance, that has been fine-tuned on oracle-agent trajectories for the train tasks.

\vspace{-0.2\baselineskip}
\textit{Oracle:} We use oracle trajectories for task-wise intervention results of Sec.~\ref{sec:self_reg_limit}. However, oracle trajectories are static and cannot be used for all settings; details are in Sec.~\ref{sec:self_reg_limit}.
\vspace{-0.5\baselineskip}

\section{Self-Regulation with PRMs and Limitations}
We first address self-regulation: Can an agent stop executing the task if it deems it cannot complete the task (Fig.~\ref{fig:fig1}(a)). 

\vspace{-0.5\baselineskip}
\subsection{Method}
\label{sec:self_reg_method}
\vspace{-0.2\baselineskip}
We measure the difficulty of a state \(s\) as \(1 - p(s)\), where \(p(s)\) is the probability of success from \(s\) up to a terminal state. To decide when to self-regulate, we train a Process Reward Model (PRM) by rolling out trajectories of base actors, following the approach of \cite{wang2024math}. Concretely, a 3B LLaMA model with a scalar head is fine-tuned (SFT) on \((\text{state}, \text{outcome})\) pairs, where the outcome is binary success or failure for the trajectory originating at that state; the PRM is trained to learn \(p(s)\). Finally, we calibrate the PRM’s threshold using a held-out validation set.

\subsection{Self-Regulation Performance} 
\label{sec:self_reg_perf}

To evaluate self-regulation performance, we first identify the maximum \((1 - \text{PRM score})\) encountered in a trajectory before the final step. This value serves as a proxy for \(1 - p(s)\) and reflects the PRM’s estimation of the most difficult state in the trajectory. We then set a threshold on this maximum \((1 - \text{PRM score})\) to optimize overall accuracy (binary success/failure) on a held-out validation set. Table~\ref{tab:self-reg} reports the accuracy, precision, and recall metrics (labeling task success as 1) on the test set for two base actors, GPT-4o-mini and LLaMA, each with its own separately trained PRM. These results demonstrate high precision and recall, indicating that the PRM score is indeed a strong indicator of \(p(s)\).

\begin{table}[!t]
\vspace{-1em}
\caption{\textbf{Self-regulation performance for base actors.}}
\setlength{\tabcolsep}{3pt} 
\centering
\begin{tabular}{l c c c c}
\toprule
 & \multicolumn{2}{c}{GPT 4o-mini} & \multicolumn{2}{c}{Llama} \\
\cmidrule(lr){2-3} \cmidrule(lr){4-5}
 & Pnp & S\_Obj & Pnp & S\_Obj  \\
\midrule
Accuracy & 90\% & 90\% & 88\% & 90\% \\
Precision & 88\% & 70\% & 88\% & 83\% \\
Recall & 100\% & 100\% & 100\% & 83\% \\
\bottomrule
\end{tabular}
\label{tab:self-reg}
\vspace{-1em}
\end{table}

\subsection{Can PRM Thresholding Work for Multi-Step Intervention Requests?}  
\label{sec:self_reg_limit}

We now investigate whether the PRM thresholding strategy, which worked for self-regulation, can also be used for multi-step intervention requests. Formally, consider a stochastic process with states \(s \in \mathcal{S}\) and transition probabilities \(P(s' \mid s)\). Each state \(s\), whether terminal or non-terminal, has a probability \(p(s)\) of eventually leading to task success. At each non-terminal state \(s\), the helper chooses either \textit{help} or \textit{nohelp}. Unlike self-regulation—where the agent simply halts and asks for help once—multi-step interventions occur at each step of the process. In practice, the intervention could come from a more capable model, an MCTS approach, or an oracle actor. The question is whether we can rely on the same measure \((1 - p(s))\) from the PRM, \textit{without} explicitly accounting for the dynamics \(P(s' \mid s)\), to identify states requiring intervention in this more granular setting.

We consider two settings:
\begin{enumerate}
\vspace{-1em}
    \item $I(s_t)$ (a.k.a. State-wise intervention): Interventions are selectively applied to states $s_t$ during the task.
    \item $I(s_t \rightarrow T)$ (a.k.a. Task-wise intervention): Once a state $s_t$ is identified, the intervention continues for the remainder of the task $T$.
    \vspace{-1em}
\end{enumerate}

Details on implementation of $I(s_t)$ and $I(s_t \rightarrow T)$ are in Appendix~\ref{app:explanation}. Table~\ref{tab:big_table} summarizes the results across different interventions, including oracle actors, larger models, and MCTS. We find the following:

\begin{enumerate}[leftmargin=1em, itemsep=3pt, parsep=0pt, topsep=2pt, partopsep=0pt]
    \item For \(I(s_t \to T)\), PRM scores consistently identify tasks that require help, aligning with self-regulation results.
    \item However, \(I(s_t)\) consistently outperforms \(I(s_t \to T)\). Under similar intervention budgets, intervening \textit{selectively at difficult states} can be more effective than intervening \textit{for all states of difficult tasks}.
    \item PRM thresholding is ineffective for \(I(s_t)\).
\end{enumerate}

\begin{figure}[!t]
    
    \centering
    \includegraphics[width=0.5\textwidth]{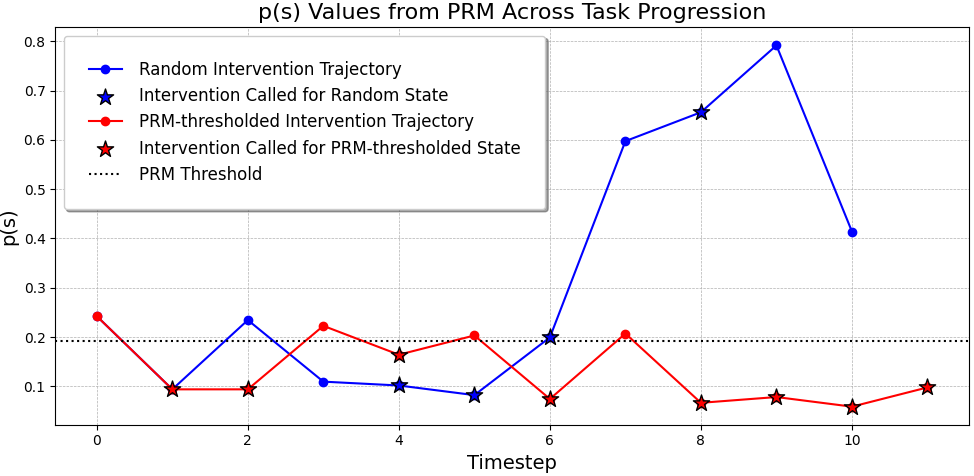}
    \vspace{-2em}
    \caption{$p(s)$ measured by the PRM across the task. Interventions on PRM-chosen states (red line and stars) cause repeated toggling that traps the agent in low-$p(s)$ regions, resulting in worse outcomes than random interventions (blue line and stars), which ends at step 10 with task success.}
    \label{fig:break}
    \vspace{-2em}
\end{figure}

We provide an in-depth discussion of these three points in Appendix~\ref{app:explanation}. Here, we highlight why PRM scores are ineffective for \(I(s_t)\). As illustrated in Figure~\ref{fig:break}, once an intervention briefly reduces the difficulty (and increases $p(s)$), the base actor immediately takes control. Before long, the difficulty surpasses the threshold again, prompting another intervention. This repeated “toggling” between the intervention and the base actor leads to suboptimal performance and illustrates the pitfalls of using only difficulty-based thresholds without accounting for sequential dependencies. Consequently, we \textit{cannot} rely on the same measure \((1 - p(s))\) used in self-regulation \emph{without} explicitly incorporating transition dynamics \(\bigl(P(s' \mid s)\bigr)\) to identify states that need intervention in this more granular setting.

\begin{figure*}[!t]
    \centering
    \includegraphics[width=1.0\textwidth]{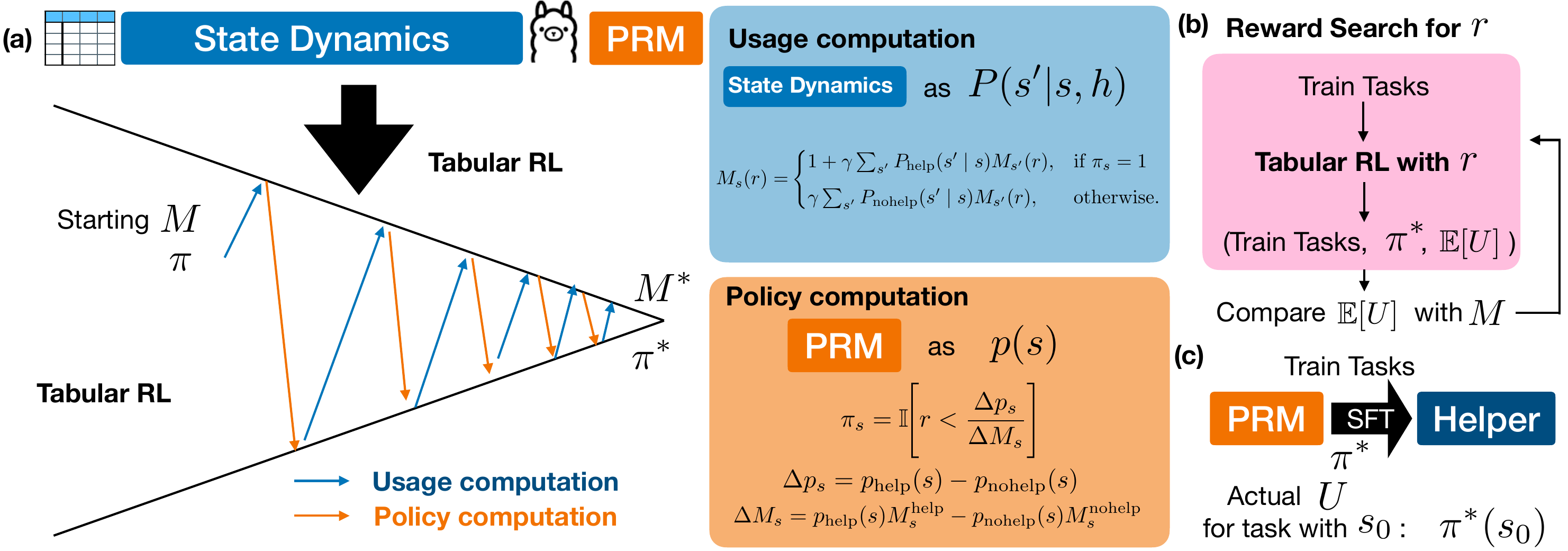}
    \vspace{-2em}
    \caption{\textbf{Method Overview.} \textbf{(a)} We combine tabular state dynamics with a \textit{process reward model} (PRM), implemented as a large language model, to perform \textit{offline} tabular reinforcement learning. Our method consists of iterative usage/policy computation. $\pi_s=\text{help}$ denoted as $\pi_s=1$ for space. \textbf{(b)} We run this offline tabular RL procedure to generate trajectory annotations for training tasks. \textbf{(c)} Finally, we train another large language model with a scalar head via supervised fine-tuning (SFT), with the trajectory annotations from step (b).}
    \label{fig:method}
    \vspace{-1em}
\end{figure*}

\vspace{-0.5\baselineskip}
\section{Method: Requesting Targeted Interventions}
\label{sec:method}

Determining \emph{when to ask for help} is challenging, especially under a constrained \emph{usage budget}. Although a PRM can estimate success probabilities, it does not capture how help requests affect future states, highlighting the need for \emph{transition dynamics} (Sec.~\ref{sec:self_reg_limit}, Fig.~\ref{fig:break}). Fully accounting for state dynamics, difficulty, usage, and help-asking policies requires searching over both rewards and policies. However, running a complete deep RL pipeline for iterative reward and policy optimization can be prohibitively expensive, requiring substantial training time, many interventions, and extensive hyperparameter tuning. In addition, collecting data for policy training is costly and constrained, prompting the question of how to effectively gather and reuse data across multiple budgets while controlling the number of interventions for collecting training data.

We propose a method that is both \textbf{offline}, requiring only a single pass of intervention data collection for training, and \textbf{hybrid}, integrating a learned PRM with classical tabular RL. This method keeps \emph{usage}---the number of times an intervention is used during inference --- under the given budget, while finding the optimal \emph{policy}---deciding whether to request help at non-terminal states. The key components of our method are as follows (Fig.~\ref{fig:method}): \emph{Transition Model Collection}, \emph{Dynamic Programming (DP) for Usage/Policy Iteration} (Fig.~\ref{fig:method}(a)), \emph{Reward Search} (Fig.~\ref{fig:method}(b)), and \emph{Final Training} (Fig.~\ref{fig:method}(c)). Their details are explained in Sec.~\ref{sec:method_alg}.

This approach offers several advantages. First, the DP step is extremely fast, completing in minutes without requiring GPUs or additional intervention requests. We do \textbf{NOT} have to train the policy itself for reward search; we only need to repeat the quick DP process. Second, the off-line nature enables adaptability across budgets. Intervention data only needs to be collected once in the transition model collection, and can be reused to obtain trajectories for different budgets. Third, using learned PRMs with tabular RL enhances robustness while avoiding the inefficiencies of deep RL. The claims on robustness are supported by results of Sec.~\ref{sec:analysis}.

\vspace{-0.5\baselineskip}
\subsection{Reward Regime and Notations}
\label{sec:method_formulation}
\vspace{-0.5\baselineskip}

Agents can request help at non-terminal states to boost success, balancing help costs. We want to maximize task success rate, while calling less than $C$ interventions in the test set. For this, we define the reward structure as follows:

\vspace{-0.2\baselineskip}
\textbf{Reward Regime}  
At any non-terminal state \(s\), the helper chooses either \emph{help} or \emph{nohelp}. 
\begin{itemize}[leftmargin=*]
    \vspace{-1em}
    \item \textbf{Intermediate states:}  
    \vspace{-0.5em}
    \begin{itemize}[leftmargin=*]
        \item \emph{help:} incurs an immediate reward of \(-r\).
        \item \emph{nohelp:} immediate reward of 0.
    \end{itemize}
    \vspace{-1em}
    \item \textbf{Terminal States:} success yields a reward of \(+1\) and failure \(0\).
    \vspace{-1em}
\end{itemize}
The cost $C$ is embedded in $r$ and we aim to find the reward $r$, together with the entailing policy, that incurs cost $C$.

\textbf{Notations} 
Success probability under \textit{help}/\textit{nohelp} are denoted

\begingroup
\vspace{-2em}
\setlength{\abovedisplayskip}{3pt}
\setlength{\belowdisplayskip}{3pt}
\setlength{\jot}{0pt}%
\begin{align*}
p_{\text{nohelp}}(s) &= \Pr(\text{success at terminal state after } s \mid \text{nohelp}), \\
p_{\text{help}}(s) &= \Pr(\text{success at terminal state after } s \mid \text{help}).
\end{align*}
\endgroup
at state $s$. Similarly, we denote state transition dynamics under \textit{help}/\textit{nohelp} at state $s$ as $P_{\mathrm{nohelp}}(s' \mid s)$, $P_{\mathrm{help}}(s' \mid s)$.

\vspace{-0.5\baselineskip}
\subsection{Derivaion of Usage/Policy Iteration}
\label{sec:method_derivation}
\vspace{-0.5\baselineskip}

\textbf{Overview } We derive the usage/policy iteration algorithm (Fig.~\ref{fig:method}(a)), effectively equivalent to value iteration, by decomposing the value function into \emph{success} and \emph{usage} components. Unlike standard value iteration, usage/policy iteration is \textit{offline} and integrates PRMs for robust classical RL.

\textbf{Summary of Usage/Policy Iteration Derivation }  
The value function \(V_s(r)\) for an optimal policy $\pi^*$ represents the expected return from state \(s\) under the reward regime:
\[
\begin{aligned}
V_s(r) 
&= \text{(expected discounted reward from \(s\),} \\
&\quad \text{with \(-r\) per help, \(+1\) at success)}.
\end{aligned}
\]
It satisfies the piecewise Bellman Equation
\[
V_s(r) =
\begin{cases}
\gamma \sum_{s'} P_{\mathrm{nohelp}}(s' \mid s) V_{s'}(r), & \text{if nohelp at $s$}, \\[6pt]
-r + \gamma \sum_{s'} P_{\mathrm{help}}(s' \mid s) V_{s'}(r), & \text{if help at $s$},
\end{cases}
\]
with \(V_{s_{\text{term}}}(r) = 1\) at terminal success states and   \(V_{s_{\text{term}}}(r) = 0\) at terminal failure states. We can disentangle $V_s$ into success and usage components:
\vspace{-0.2\baselineskip}
\begin{center}
\(
V_s(r)
\;=\;
S_s
\;-\;
r\,M_s(r), \quad \text{where}
\)
\end{center}
\vspace{-1em}
\begin{center}
\(
S_s
= 
\mathbb{E}_{\pi^*}\Bigl[\sum_{t=0}^{\infty} \gamma^t\,1_{\text{success at time }t}\Bigr]
\)
\(
M_s(r)
= 
\mathbb{E}_{\pi^*}\Bigl[\sum_{t=0}^{\infty} \gamma^t\,1_{\text{help at time }t}\Bigr]
\)
\vspace{-1em}
\end{center}

This decomposition, together with the piecewise Bellman equation, leads us to the recursive definition of optimal usage and policy.

\begingroup
\centering
\setlength{\abovedisplayskip}{0pt}
\setlength{\abovedisplayshortskip}{0pt}
\setlength{\belowdisplayskip}{0pt}
\thinmuskip=1mu
\medmuskip=1mu
\thickmuskip=1mu
\relax
\vspace{-1em}
\[
\resizebox{1.0\columnwidth}{!}{%
\boxed{%
\begin{aligned}
M_s(r)=&
  \begin{cases}
    M_s^{\mathrm{help}}=1+\gamma\sum_{s'}P_{\mathrm{help}}(s'\mid s)M_{s'}(r),
      & \text{if }\pi(s)=\text{help},\\[4pt]
    M_s^{\mathrm{nohelp}}=\gamma\sum_{s'}P_{\mathrm{nohelp}}(s'\mid s)M_{s'}(r),
      & \text{otherwise}.
  \end{cases}\\[4pt]
\pi(s)=&\text{help} \;\iff\; r<\frac{\Delta p_s}{\Delta M_s}.
\end{aligned}%
}%
}
\]
\endgroup

where we denote \(\Delta p_s = p_{\text{help}}(s) - p_{\text{nohelp}}(s)\) and 
\begin{center}
\vspace{-0.5em}
\(
\Delta M_s =  p_{\text{help}}(s) M_s^{\text{help}} -p_{\text{nohelp}}(s)  M_s^{\text{nohelp}}.
\)
\end{center}
Usage/policy iteration (the boxed equation) is a direct consequence of (i) substituting \(V_s(r) = S_s - r\,M_s(r)\) into the Bellman recursion and (ii) deriving and applying the threshold condition of $r$ for comparing help vs.\ nohelp. Please see Appendix~\ref{app:derivation} for more detailed derivation.

\vspace{-0.5em}
\textbf{Consequences and Explanations } Iterating the boxed equation converges to a unique fixed point \(M^*(r)\) (Proof in Appendix~\ref{app:convergence}). The optimal DP policy $\pi^*(r)$ is then obtained by determining which branch each state \(s\) selects in the stable solution: \emph{help} if \(\Delta p_s / \Delta M^*_s > r\), and \emph{nohelp} otherwise. Intuitively, \emph{help} is optimal precisely if 
\begin{center}
\vspace{-0.5em}
\(
\text{(extra success)} 
\;>\; 
r \text{(extra cost)} 
\Longleftrightarrow
\Delta p_s 
\;>\;
r \;\Delta M^*_s
\)
\end{center}

\vspace{-0.75em}
\subsection{Algorithm}
\label{sec:method_alg}

An overview of the algorithm to compute the fixed point \( M^*(r) \) and \( \pi^*(r) \) is provided in Fig.~\ref{fig:method}. The process begins with collecting a table of transition dynamics. Using these dynamics and a learned PRM, the algorithm iteratively alternates between reward search (Fig.~\ref{fig:method}(b)) and usage/policy search (Fig.~\ref{fig:method}(a)) to converge on the desired solution.

\textbf{Phase 1: Exploration and Building Transition Model }
We collect transitions by randomly triggering interventions using the base actor on training tasks. For each transition, we update the count as follows:
\begin{center}
\(
\texttt{count}[s][a][s']{+}{+}.
\)
\end{center}
Specifically, we perform three randomly seeded repetitions of rollouts where interventions are triggered with probabilities of 0.0, 0.1, 0.3, 0.5, 0.7, 0.9, and 1.0 for each task. For scenarios involving multiple interventions (see Table~\ref{tab:main}), we collect transitions for each intervention individually and additionally for combinations of intervention probabilities, with 0.1/0.1, 0.3/0.3, 0.1/0.3, and 0.3/0.1 per intervention. After exploration, raw counts are normalized to estimate the transition probabilities, where $a \in \{\text{help},\text{nohelp}\}$.
\begin{center}
\(
\hat{P}(s'\mid s,a) = \frac{\texttt{count}[s][a][s']}{\sum_x \texttt{count}[s][a][x]}.
\)
\end{center}

\textbf{Phase 2: Offline DP for Iterative Reward/Policy Search}
We seek a cost parameter \(r\) that meets usage budget constraint $C$ (Sec.~\ref{sec:method_formulation}). In practice, we can systematically \emph{search} over possible \(r\) values (e.g.\ via binary search) and, for each candidate \(r\), run a fast offline DP to obtain both the expected usage \(\mathbb{E}[U]\) and the associated policy.

Concretely, for a given \(r\) (see Fig.~\ref{fig:method}), we initialize usage estimates \( M^{\text{help}}_s,\, M^{\text{nohelp}}_s \) and iterate the boxed equation of Sec.~\ref{sec:method_derivation}. We then compare \(\mathbb{E}[U] = M_{s_0}\) with the usage budget or other constraints, and adjust \(r\) as necessary.  Since each iteration is purely tabular, it is very fast and does not require new environment rollouts.  

\noindent \textbf{Initialization.}  
Set \(M_s^{(0)} = 0\) (or any other guess) for all states \(s \in \mathcal{S}\) collected in Phase 1, and set \(\texttt{iteration} = 0\).

\textbf{Usage Computation}
    For each state \(s\), compute
        \[
        M_s^{\text{help}} 
        \;\leftarrow\;
        1 \;+\;\gamma \sum_{s'} P_{\mathrm{help}}(s'\mid s)\,M_{s'}^{(\texttt{iteration}-1)},
        \]
        \[
        M_s^{\text{nohelp}}
        \;\leftarrow\;
        \gamma \sum_{s'} P_{\mathrm{nohelp}}(s'\mid s)\,M_{s'}^{(\texttt{iteration}-1)}.
        \]
\textbf{Policy Computation} First, compute $\Delta p_s, \Delta M_s$ as defined in Sec.~\ref{sec:method_derivation}. If \(\Delta M_s\) is (numerically) very close to zero, default to \(\texttt{nohelp}\).  Otherwise, 
\[
\text{ratio} \;=\; \frac{\Delta p_s}{\Delta M_s}, 
M_s^{(\texttt{iteration})} \;=\;
\begin{cases}
M_s^{\text{help}}, & \text{if } r < \text{ratio} \\[4pt]
M_s^{\text{nohelp}}, & \text{otherwise}.
\end{cases}
\vspace{-0.3em}
\]

We iteratively repeat usage computation/policy computation for \(\texttt{iteration} = 1,2,\dots,\texttt{max\_iters}\) or until we hit the convergence critertion: \(\Delta_{\max} < \varepsilon\) (the tolerance), where 
\begin{center}
\(
\Delta_{\max}
\;=\;
\max_{s \in \mathcal{S}}
\bigl|
M_s^{(\texttt{iteration})}
\;-\;
M_s^{(\texttt{iteration}-1)}
\bigr|.
\)
\end{center}

We can obtain the following outcomes from the converged $M_s(r)$. First, the DP policy $\pi^*(r)$ is then:
\begin{center}
\(
\pi_s^*(r) 
\;=\;
\begin{cases}
 \texttt{help}, & \text{if } M_s(r) = M_s^{\text{help}},\\[3pt]
 \texttt{nohelp}, & \text{otherwise}.
\end{cases}\)
\end{center}

The expected usage from the start state \(s_0\) is \(\mathbb{E}[U] = M_{s_0}(r)\). We compare \(\mathbb{E}[U]\) to the usage budget $C$ and, if unsatisfactory, adjust \(r\) and re-run the above offline DP.  This approach supports quick \emph{re-solving} for multiple \(r\) values without further data collection.

\textbf{Phase 3: Final Policy Training via SFT }  
After deriving tabular $\pi^*(r)$ for all states \(s \in \mathcal{S}\) collected in Phase 1, we train the helper model using standard supervised finetuning (SFT).  Concretely, the helper model learns to replicate the DP policy $\pi^*(r)$'s help/nohelp decisions \textit{from the train tasks} for downstream deployment. This helper is plugged in as in Fig.~\ref{fig:sif}(b) and is evaluated on the test set.

\vspace{-1em}
\subsection{Extension to Multiple Interventions}
\label{sec:method_multi}
Our algorithm extends to multiple intervention types, each with its own budget. The same framework in Sections \ref{sec:method_derivation}/~\ref{sec:method_alg} can be applied, with adaptations to handle multiple interventions with individual budget constraints, each with a different cost and expected usage (e.g. $r_1$, $M_s^1$ for \textit{help1} and $r_2$, $M_s^2$ for \textit{help2}). In short, we can adapt Phase 2 to select the action with minimal combined cost \(\;r_1\,M_s^1 + r_2\,M_s^2\). Details are in Appendix~\ref{app:multiple} and results are in Tab.~\ref{tab:main}.

\begin{table}[!t]
\vspace{-1em}
\caption{Performance and intervention usage comparison of our method and baselines, across task types. A more powerful model was used as the intervention.}
\centering
\normalsize
\setlength{\tabcolsep}{3pt}
\resizebox{\columnwidth}{!}{%
\begin{tabular}{l c c c c c c c c c c}
\toprule
& \multicolumn{5}{c}{\textbf{S\_Obj}} 
& \multicolumn{5}{c}{\textbf{Pick N Place}} \\
\cmidrule(lr){2-6}\cmidrule(lr){7-11}
& SR \(\uparrow\) & SPL \(\uparrow\) & \( L \downarrow \) & \( U \downarrow \) & \(\mathbb{E}[U]\) 
& SR \(\uparrow\) & SPL \(\uparrow\) & \( L \downarrow \) & \( U \downarrow \) & \(\mathbb{E}[U]\)  \\
\midrule
0\% Interv. & 30.0 & 26.6 & 12.3 & 0.0 & - & 35.0 & 30.1 & 8.0 & 0.0 &  - \\
100\% Interv. & 67.5 & 65 & 7.8 & 7.8 & - & 60.0 & 52.0 & 4.6 & 4.6 &  - \\
\midrule
\rowcolor{gray!40}
\multicolumn{11}{l}{Random} \\
10\% & 47.5 & 38 & 10.5 & 1.1 & - & 37.5 & 32.9 & 6.8 & 0.6 &  - \\
30\% & 50 & 44.1 & 8.9 & 2.9 & - & 50 & 42.6 & 5.8 & 1.5 &  - \\
\midrule
\rowcolor{gray!40}
\multicolumn{11}{l}{State-wise PRM Thresholding}  \\
10\% & 42.5 & 35.5 & 11.8 & 1.0 & - & 32.5 & 27.1 & 7.0 & 0.6 & - \\
30\% & 42.5 & 35.5 & 11.1 & 1.8 & - & 57.5 & 39.3 & 4.8 & 3.2 & - \\
\midrule
\rowcolor{gray!40}
\multicolumn{11}{l}{\textbf{Our Method}} \\
\(r\) high & 47.5 & 39.3 & 11.4 & 0.4 & 0.4 & 47.5 & 36.6 & 6.0 & 1.2 & 0.7\\
\(r\) mid  & \textbf{62.5} & \textbf{57.5} & \textbf{9.1} & \textbf{1.0} & \textbf{0.8} & 60.0 & 49.7 & 5.7 & 1.6 & 0.9\\
\(r\) low  & 60.0 & 54.5 & 8.7 & 2.2 & 1.1 & \textbf{64.9} & \textbf{58.9} & \textbf{5.2} & \textbf{2.9} & \textbf{1.8} \\
\bottomrule
\end{tabular}%
}
\vspace{-2em}
\label{tab:result_task}
\end{table}
\vspace{-1em}
\section{Results}
We provide results for our method, with the LLaMa base actor and better model/MCTS interventions. We plug in our trained helper from Sec.~\ref{sec:method}, as in Fig.~\ref{fig:sif}(b), to call intervention(s) when the helper chooses \textit{help}. As baselines, we compare to triggering intervention with random state selection and the PRM state selection in Sec.~\ref{sec:self_reg_limit}.

\begin{table*}[!ht]
\vspace{-1em}
\caption{Performance and intervention usage comparison of our method and baselines, across intervention types, evaluated on S\_obj tasks. Our method achieves performance close to using intervention 100\%, while using much fewer interventions.}
\centering
\normalsize
\setlength{\tabcolsep}{3pt}
\resizebox{1.0\textwidth}{!}{%
\begin{tabular}{l c c c c c c c c c c c c c c c c c c }
\toprule
& \multicolumn{5}{c}{\textbf{A More Powerful Model}}
& \multicolumn{5}{c}{\textbf{MCTS}}
& 
& \multicolumn{7}{c}{\textbf{Multiple Interventions}} \\
\cmidrule(lr){2-6}\cmidrule(lr){7-11}\cmidrule(lr){13-19}
& SR \(\uparrow\)
& SPL \(\uparrow\)
& \( L \downarrow\ \)
& \( U \downarrow\)
& \(\mathbb{E}[U]\)
& SR \(\uparrow\)
& SPL \(\uparrow\)
& \( L \downarrow\ \)
& \( U \downarrow\)
& \(\mathbb{E}[U]\)
& 
& SR \(\uparrow\)
& SPL \(\uparrow\)
& \( L \downarrow\ \)
& \( U_1 \downarrow\)
& \( U_2 \downarrow\)
& \(\mathbb{E}[U_1]\) 
& \(\mathbb{E}[U_2]\)\\
\midrule

0\% Interv.
& 30 & 27 & 12.3 & 0 & --
& 30 & 27 & 12.3 & 0 & --
& 0\% Interv.
& 30 & 27 & 12.3 & 0 & -- & -- & --  \\

100\% Interv.
& 68 & 65 & 7.8 & 7.8 & --
& 63 & 52 & 9.4 & 9.4 & --
& --
& -- & -- & -- & -- & -- & -- & -- \\

\midrule
\rowcolor{gray!20}
\multicolumn{19}{l}{\textbf{Random}} \\

10\%
& 48 & 38 & 10.5 & 1.1 & --
& 43 & 37 & 11.3 & 1.4 & --
& 10\%, 10\%
& 43 & 33 & 11.3 & 1.1 & 0.8 & -- & -- \\

30\%
& 50 & 44 & 8.9 & 2.9 & --
& 48 & 38 & 11.2 & 3.7 & --
& 30\%, 30\%
& 48 & 40 & 9.9 & 3.2 & 2.6 & -- & -- \\

\midrule
\rowcolor{gray!20}
\multicolumn{19}{l}{\textbf{State-wise PRM Thresholding}} \\

10\%
& 43 & 36 & 11.8 & 1.0 & --
& 40 & 31 & 12.1 & 1.1 & --
& 20\% \& random
& 40 & 29 & 12.4 & 0.6 &  0.9 & -- & --  \\

30\%
& 43 & 36 & 11.1 & 1.8 & --
& 38 & 30 & 11.6 & 2.5 & --
& 50\% \& random
& 48 & 39 & 10.8 & 2.0 &  2.2 & -- & \\

\midrule
\rowcolor{gray!20}
\multicolumn{19}{l}{\textbf{Our Method}} \\

\(r\) high
& 48 & 39 & 11.4 & 0.4 & 0.4
& 38 & 27 & 12.3 & 0.5 & 0.7
& \hspace{0.5em} \(r_1\) high, \(r_2\) high 
& 38	  & 34 & 11.5 & 0.1	 & 1.2 & 0.3  & 0.8   \\

\(r\) mid
& \textbf{63} & \textbf{58} & \textbf{9.1} & \textbf{1.0} & \textbf{0.8}
& 45 & 37 & 11.6 & 1.4 & 1.0
& \hspace{0.5em} \(r_1\) high, \(r_2\) mid
& 43 & 35 & 10.7 & 0.1 & 1.8 & 0.4  & 1.0  \\

\(r\) low
& 60 & 55 & 8.7 & 2.2 & 1.1
& \textbf{50} & \textbf{36} & \textbf{10.3} & \textbf{3.2} & \textbf{1.3}
& \hspace{0.5em} \(r_1\) mid, \(r_2\) high
& \textbf{48} & \textbf{42} & \textbf{9.9} & \textbf{2.0} & \textbf{0.6} & \textbf{1.0} & \textbf{0.7}  \\

\bottomrule
\end{tabular}%
}
\label{tab:main}
\vspace{-1em}
\end{table*}
\vspace{-0.5\baselineskip}
\subsection{Main Results}

\textbf{Results across tasks }
Table~\ref{tab:result_task} compares our method to baselines in terms of success rate (\(\mathrm{SR}\)), path-length weighted success (\(\mathrm{SPL}\)), task execution length (\(L\)), observed intervention usage (\(U\)), and expected intervention usage (\(\mathbb{E}[U]\)). We compute \(\mathbb{E}[U]\) by averaging \(M(s_0)\) for starting states \(s_0\) under \(\pi^*\). Note that  \(\mathbb{E}[U]\) is only applicable to our method, since it is not straightforward to know this for other methods. We train our approach using different reward scale values (\(r\) high, mid, low), inducing varying intervention frequencies. 

With just a fraction of the interventions used by a policy that always intervenes (7.8 and 4.5 times on average), our method nearly matches that policy’s performance. For example, in \emph{S\_obj}, we achieve a \(62.5\%\) success rate using only \(1.0\) intervention on average, outperforming baselines with similar or higher usage. Moreover, \(\mathbb{E}[U]\) closely matches observed usage, especially for smaller \(U\) (e.g. $U$ is 0.4 and $\mathbb{E}[U]$ is also 0.4). They tend to diverge more with higher $r$'s, but \(\mathbb{E}[U]\) still provides good expectations of the model's intervention usage, allowing us to select \(r\) based on training data alone, without exhaustive training and evaluation.

\begin{figure}[!t]
    \centering
    \includegraphics[width=1.0\columnwidth]{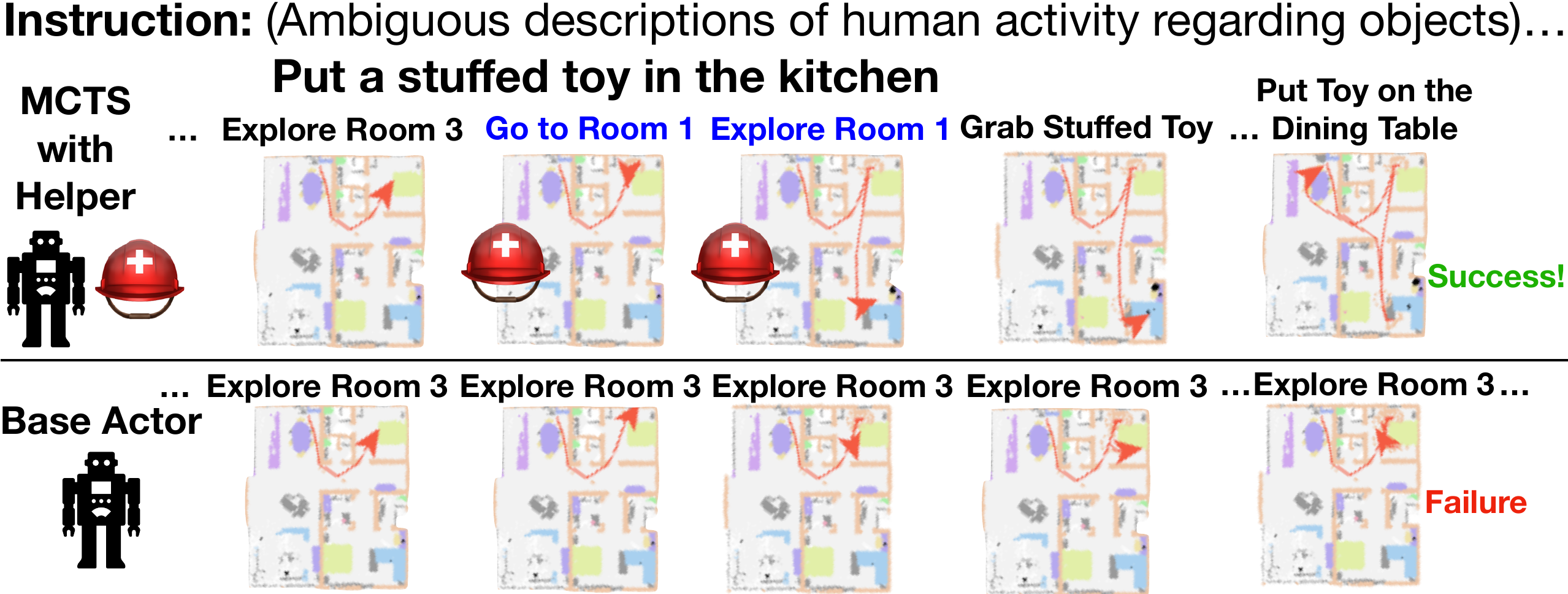}
    \vspace{-2em}
    \caption{Example Trajectory with interventions and base actor.}
    \label{fig:example}
    \vspace{-2em}
\end{figure}

We find that performance drops at \textit{\(r\)~high} are driven by the base LLaMA actor, not our method. With frequent interventions, the base actor encounters more out-of-distribution (OOD) states and produces invalid actions. Because the base actor was trained only on its own trajectories, these new, intervention-driven states lie outside its familiar distribution and makes the base actor produce invalid actions (for instance, actions not in the allowed set). This explains about 75\% of failures at \textit{\(r\)~high}, but under 5\% at \textit{\(r\)~low} and 15\% at \textit{\(r\)~mid}. We do not observe these anomalies for \emph{PnP} or other interventions (Table~\ref{tab:main}), likely because the strategy employed by MCTS or the stronger intervention model remains closer to the base actor’s distribution in those scenarios. Moreover, this behavior does not appear in training tasks, where the success rate at \textit{\(r\)~high} exceeds that at \textit{\(r\)~mid} or \textit{\(r\)~low} (Tab.~\ref{tab:analysis}). A likely reason is that repeated interventions on the same tasks the base actor was trained on keeps it in the distribution that the base actor has effectively memorized.

\textbf{Results across interventions } 
Table~\ref{tab:main} compares our method and baselines on the more challenging \emph{S\_obj} split, evaluating three intervention setups: a better model, MCTS, or both. For multiple interventions, we use the Phase~3 extension from Sec.~\ref{sec:method_multi} to assign individual budgets. Consequently, we present results with different \((r_1, r_2)\) configurations, where \(r_1\) controls usage of the better model and \(r_2\) controls MCTS. For baselines for multiple interventions, we randomly select states (10\% or 30\%) for each intervention, resulting, for example, in 10\%,10\% and 30\%,30\%. In the state‐wise PRM thresholding baseline, we calibrate thresholds for 20\% and 50\% of states and trigger each intervention randomly half of the times. For single interventions, we follow the same protocols as in Table~\ref{tab:result_task}. 

In general, the trends from Table~\ref{tab:result_task} hold here as well. First, our method optimally calls interventions, whether MCTS or both, achieving higher performance than baselines while using fewer interventions (e.g., with only 0.5 MCTS calls on average, we match the success rate of a 30\% PRM thresholding baseline that uses 2.5 calls). Second, using multiple interventions does not yield substantially better results than a single intervention under similar usage constraints, likely due to strategy clashes (Fig.~\ref{fig:break}). Nevertheless, our method still outperforms multi-intervention baselines. Finally, we find that \(\mathbb{E}[U]\) remains a reliable predictor of actual usage \(U\), especially at higher \(r\) values, providing a useful guide for choosing budgets in advance.

\begin{figure}[!t]
    \centering
    \includegraphics[width=0.5\textwidth]{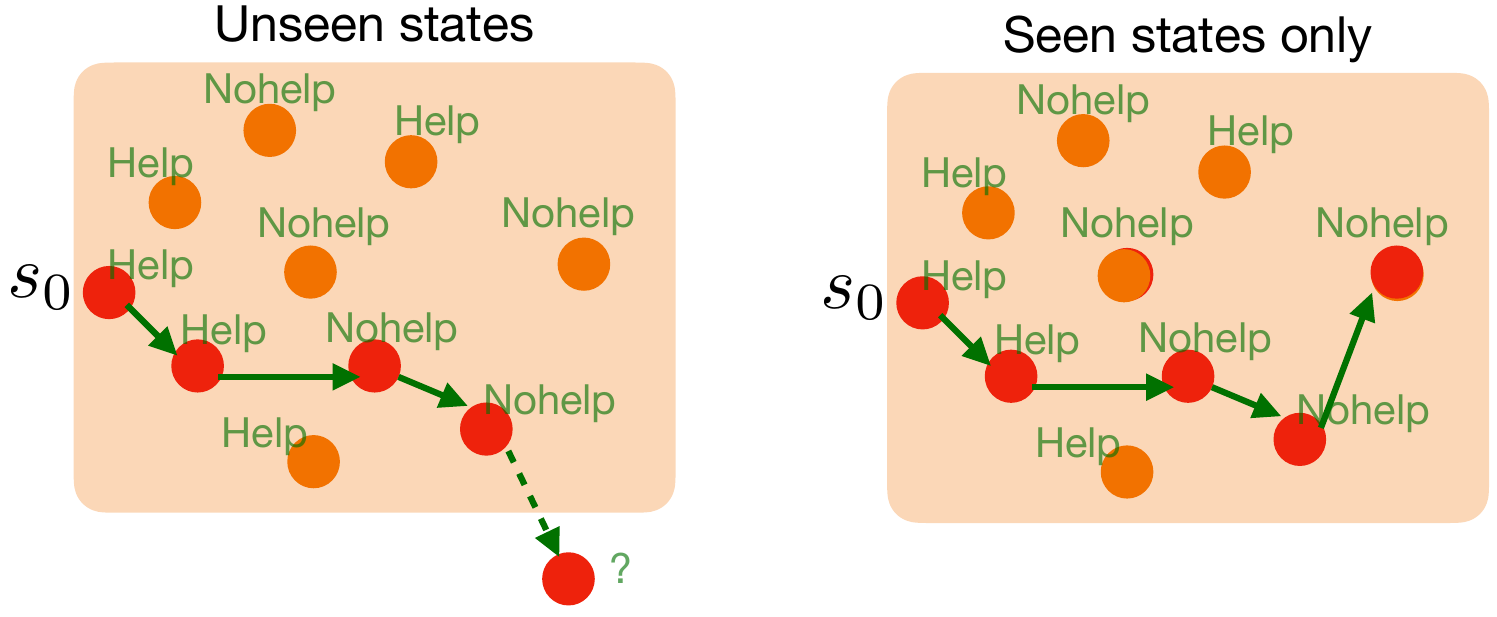}
    \vspace{-2em}
    \caption{Seen vs.\ unseen states in the training data of the helper. The orange region highlights all states collected in Phase~I (Sec.~\ref{sec:method_alg}), each labeled with $\pi^*$ (\emph{Nohelp} or \emph{Help} action). The green arrow illustrates $\pi^*$ rollout from the initial state \(s_0\).}
    \label{fig:unseen}
    \vspace{-2em}
\end{figure}

\textbf{Qualitative example }  
Figure~\ref{fig:example} shows an \emph{S\_obj} task execution comparing the base actor to our helper approach, which applies MCTS-based interventions with \(r=0.5\) (see Table~\ref{tab:main}). The primary challenge is locating a stuffed toy in a cluttered environment, where ambiguous descriptions can point to multiple potential locations. The base actor begins by visiting different rooms but soon becomes stuck, repeatedly exploring Room~3. In contrast, our helper detects this stall point and calls for just two well-timed interventions, enabling a shift toward a more effective strategy and resulting in successful task completion.

\vspace{-0.5\baselineskip}
\subsection{Analysis}
\label{sec:analysis}
A key concern of off-line and tabular state collection is coverage and robustness to unseen states. Table~\ref{tab:analysis} investigates how different training data selections (selecting different outcomes from the DP process) influence the helper model’s performance and intervention usage. We compare two strategies for using the outcomes of the DP (Phase 2) as training data for the helper (Fig.~\ref{fig:unseen}):

\vspace{-0.2\baselineskip}
\textbf{All States} -- Includes every $(s,a)$ pair of $\pi^*$, for all $s$ collected in Phase~1 (the primary approach in Tables~\ref{tab:main} and~\ref{tab:result_task}). 

\vspace{-0.2\baselineskip}
\textbf{Trajectory Only} -- Starting from $s_0$, follow $\pi^*$ (green arrow of Fig.~\ref{fig:unseen}) and use the $(s,a)$ pair from this trajectory (the red states only); if $\pi^*$ encounters an unseen state, do not include this task/trajectory for training the helper.

\vspace{-0.2\baselineskip}
We then evaluate them under splits of the train set (Fig.~\ref{fig:unseen}):

\vspace{-0.2\baselineskip}
\textbf{Seen tasks} -- The task terminates following $\pi^*$.

\vspace{-0.2\baselineskip}
\textbf{Unseen tasks} -- Unseen state encountered while rolling $\pi^*$.

\vspace{-0.2\baselineskip}
Note that we do not have $\pi^*$ on val/test sets (since DP searches for trajectories), and this splitting is only applicable to train tasks. Under \textbf{Trajectory Only} method, the helper policy struggles in Unseen tasks -- showing high intervention usage $(U)$, low success rates (SR), and a large discrepancy between realized and expected usage (e.g., $5.4$ vs.\ $1.14$). By contrast, \textbf{All States} maintains better alignment between $U$ and $\mathbb{E}[U]$ alongside higher SR. Broader sampling in Phase~I could further improve performance for Unseen tasks.

\begin{table}[!t]
\vspace{-1em}
\caption{Performance and Intervention Usage Comparison (\emph{S\_obj} only; SPL columns omitted).}  
\centering  
\normalsize  
\setlength{\tabcolsep}{3pt}  
\resizebox{\columnwidth}{!}{%
\begin{tabular}{l c c  c c c c c }  
\toprule  
& \multicolumn{2}{c}{\textbf{Unseen}}  
& \multicolumn{2}{c}{\textbf{Seen}}  
& \multirow{2}{*}{\textbf{Exp Usage}} 
& \multicolumn{2}{c}{\textbf{Overall}}  \\
\cmidrule(lr){2-3}\cmidrule(lr){4-5}\cmidrule(lr){7-8}
& SR & Usage
& SR & Usage
& 
& Train SR & Test SR \\  
\midrule  
\rowcolor{gray!40}  
\multicolumn{8}{l}{\textbf{Our Method (All States)}} \\  
$r$ high & 26 & 0.25 &  55  & 0.46 & 0.49 & 46 & 48 \\  
$r$ mid  & 49 & 0.78  & 63 & 0.92 & 0.43 & 56 & 63  \\  
$r$ low  & 52 & 1.81 & 60 & 1.86 & 0.92 & 60 & 60 \\  
\midrule  
\rowcolor{gray!40}  
\multicolumn{8}{l}{\textbf{Trained on Trajectory Only}} \\  
$r$ high & 16 & 1.82 & 62 & 2.19 & 0.18 & 43 & 43 \\  
$r$ mid  & 29 & 3.35 & 60 & 3.33 & 0.45 & 48 & 50 \\  
$r$ low  & 37 & 5.86 & 64 & 4.92 & 0.76 & 54 & 60 \\  
\bottomrule  
\end{tabular}%
}  
\vspace{-2em}
\label{tab:analysis}
\end{table}

\vspace{-0.5\baselineskip}
\section{Conclusion}
\vspace{-0.5\baselineskip}
We introduce an offline framework that uses classical RL combined with LLM-based PRMs to train a helper model for targeted intervention requests. By collecting transition probabilities offline and annotating tasks with optimal help-seeking strategies, our approach effectively balances performance and cost, as demonstrated in Situated Instruction Following tasks. These results show a step towards LLM agents that can self-regulate and request for interventions.

\section*{Impact Statement}

This paper focuses on enhancing the reliability of large language model (LLM) agents, with particular attention to their ability to self-regulate and ask for targeted interventions. This paper has potential societal implications regarding safety and reliability of LLM agents. Beyond these considerations, we have not identified any additional consequences that require further discussion.

\bibliography{example_paper}
\bibliographystyle{icml2024}

\appendix 

\section{MCTS Implementation Details}
\label{app:mcts}

\[ UCT = Q(s,a) + c \times \sqrt{\frac{\ln N(s)}{N(s,a)}} \]

We train a separate PRM to get $Q(s,a)$ (the PRM in Sec.~\ref{sec:self_reg_method} takes only $s$ as input) to judge the compatibility of action $a$ with state $s$. The UCT score $c \times \sqrt{\frac{\ln N(s)}{N(s,a)}}$ is calculated from $N(s)$ and $N(s,a)$ that count the number of times state $s$ has been visited and $(s, a)$ has been proposed in the current task. 

We propose five actions $a$ from a the base actor but with temperature 1.0 (we use 0.0 without MCTS), and we use $c=0.25$. In steps where MCTS is not used (in case wehre it is applied to  selected states as in Tab.~\ref{tab:main}), we update the chosen $N(s)$ and $N(s,a)$ by multiplying the chosen count by five times.

\section{Detailed Explanation of Section 4.3}
\label{app:explanation}

In this section, we provide a thorough examination of the claims made in Sec.~\ref{sec:self_reg_limit} regarding the limitations of using PRM thresholding
for multi-step intervention requests. We reference the results reported in 
Table~\ref{tab:big_table} to support each claim.

\subsection{Implementation of $I(s_t \rightarrow T)$ and $I(s_t)$}
To identify states or tasks requiring intervention, we examine two approaches:

\textbf{PRM-based thresholding } For $I(s_t)$, in a held-out val set, states are ranked by PRM scores, and the threshold for top 10\%/30\% most difficult states are selected. For $I(s_t \rightarrow T)$, we evaluate two strategies:
\begin{enumerate}[label=(\alph*)]
    \vspace{-1em}
    \item $I(s_{[0:5]} \rightarrow T)$ (a.k.a. First five steps): A PRM score threshold is thresholded on val set, based on the initial five steps of the task. On test set, if the threshold is met, intervention is applied for the entire task onward from the sixth step.
    \item $I(s_{[0:-1]} \rightarrow T)$  (a.k.a. all steps): The base actor completes the task to determine whether intervention is needed. If so, the task is restarted, and intervention is applied from $s_0$.
    \vspace{-1em}
\end{enumerate}

\textbf{Random selection } States $s_t$ or tasks $T$ are selected randomly as a baseline for comparison.

\subsection{Interperting Table~\ref{tab:big_table}}
Table~\ref{tab:big_table} shows the success rate (SR), success-per-length (SPL), and
intervention usage rates (\%) across several intervention schemes, including both
\emph{task-wise} (top gray-shaded block) and \emph{state-wise} (bottom gray-shaded block)
strategies. The table is subdivided into three overarching intervention types: oracle (columns 2--5), more powerful model delegation (columns 6--11), and MCTS (columns 12--17). 

Each intervention type has further subdivisions for 
\textbf{task-wise} (rows labeled ``Random \%'' or 
``PRM-Thresholded: All steps'' / ``PRM-Thresholded: First five steps'' under the 
\emph{Task-wise Intervention} header) 
and 
\textbf{state-wise} (rows labeled ``Random \%'' or 
``PRM-Thresholded'' under the \emph{State-wise Intervention} header).

\vspace{0.5em}

Note that oracle trajectories are not applicable to state-wise interventions, since they are static trajectories (not models) and cannot adapt to new states. 
By contrast, the \emph{A More Powerful Model} and \emph{MCTS} columns 
(6--11 and 12--17, respectively) apply to both task-wise and state-wise triggers,
thereby showing how multi-step or partial usage of these interventions fares.

\subsection{Explanation of Section 4.3}
We reference Tab.\ref{tab:big_table} and explain how PRM scores can identify difficult tasks but struggle with sequential dependence. Below are the claims in Sec.~\ref{sec:self_reg_limit} and their evidence.

\paragraph{1) PRM scores reliably identify tasks needing assistance}
Looking at Table~\ref{tab:big_table} under \emph{Task-wise Intervention}, we see that 
\emph{PRM-Thresholded} usage (rows titled ``PRM-Thresholded: All steps'' or 
``PRM-Thresholded: First five steps'') frequently achieves higher or comparable 
SR relative to \emph{Random} usage with similar or lower 
Usage. For example, under the \emph{Oracle} columns for GPT4o mini 
(columns 2--3), 
\begin{itemize}
    \item \textbf{Random 30\%} obtains SR = 38\% and SPL = 28\%.
    \item \textbf{PRM-Thresholded: All steps 30\%} achieves SR = 75\% with SPL = 58\%.
\end{itemize}

Here, using a PRM threshold nearly doubles the SR 
(75\% vs.\ 38\%). The trend persists in other settings under \emph{Oracle} as well.
 
\paragraph{2) State-wise interventions typically outperform task-wise methods.}
Comparing the \textbf{Task-wise Intervention} block (rows 8--18 in the table) with the
\textbf{State-wise Intervention} block (rows 20--25), even random state-wise triggers
often achieve higher SR for the same or lower usage. For instance, under
\emph{A More Powerful Model} for GPT4o mini (columns 6--8):

\begin{itemize}
    \item \emph{Task-wise Random 30\%} achieves SR = 38\% and SPL 28\% with usage = 2.8.
    \item \emph{Task-wise PRM-Thresholded all steps 10\%} achieves SR = 45\% and SPL =33\% with usage = 4.1.
    \item \emph{Task-wise PRM-Thresholded first five steps 10\%} achieves SR = 28\% and SPL 19\% with usage = 1.4.
    \item \emph{State-wise Random 10\%} yields 
    SR = 48\% and SPL=33\% with usage = 1.2\%.
\end{itemize}

As we see, state-wise intervention yields the best SR and SPL, compared to task-wise interventions with more usage. 
This trend frequently persists throughout Table~\ref{tab:big_table} and it underscores how \emph{intervening selectively at 
difficult states} can be more effective than intervening for the remainder of an
entire task.

\paragraph{3) PRM-based selection can underperform random selection due to transition dynamics (the ``toggling'' issue).}
Despite PRM-based triggers being more principled for picking out tough states, 
Table~\ref{tab:big_table} shows examples where \emph{PRM-Thresholded} with a low 
percentage budget does (much) worse than \emph{Random}.
Focusing on the A More Powerful Model columns for GPT4o mini (columns~6--8):

\begin{itemize}
    \item At 10\% budget, \textbf{Random 10\%} yields 
          SR = 48\% and SPL=33\%  for intervention usage = 1.2.
    \item At 30\% budget, \textbf{PRM-Thresholded 30\%} has SR = 45\% and SPL=26\%, which is lower, at usage = 2.5.
\end{itemize}

In this example, one can see that PRM-thresholded, despite involving learning, uses much more usage than Random, while leading to low performance. Table~\ref{tab:big_table} shows that PRM-thresholded frequently lags behind random across various interventions and base actors.

The text in the main paper (Sec.~\ref{sec:self_reg_limit} and Appendix~\ref{app:explanation})
explains that once an intervention lowers the difficulty for the next step, 
the PRM score may immediately drop below the threshold, handing control back to the 
base actor too soon. When the difficulty rises again, another intervention triggers, 
leading to repeated back-and-forth ``toggling'' that is inefficient. Such a phenomenon
is not captured by a PRM approach that only scores state difficulty in isolation, 
thus harming performance in multi-step scenarios.

\begin{table*}[!t]
\caption{Task-wise and state-wise interventions performances.}
\centering
\setlength{\tabcolsep}{3pt} 
\resizebox{\textwidth}{!}{%
\begin{tabular}{l c c c c c c c c c c c c c c c c c}
\toprule
 & \multicolumn{4}{c}{\textbf{Oracle}} & \multicolumn{6}{c}{\textbf{A More Powerful Model}} & \multicolumn{6}{c}{\textbf{MCTS}} \\
\cmidrule(lr){2-5}\cmidrule(lr){6-11}\cmidrule(lr){12-17}
 & \multicolumn{2}{c}{GPT4o mini} & \multicolumn{2}{c}{SFT-ed Llama} 
 & \multicolumn{3}{c}{GPT4o mini} & \multicolumn{3}{c}{SFT-ed Llama}  
  & \multicolumn{3}{c}{GPT4o mini} & \multicolumn{3}{c}{SFT-ed Llama} \\
\cmidrule(lr){2-3}\cmidrule(lr){4-5}
\cmidrule(lr){6-8}\cmidrule(lr){9-11}
\cmidrule(lr){12-14}\cmidrule(lr){15-17}
 & SR  \(\uparrow\) & SPL  \(\uparrow\) & SR  \(\uparrow\) & SPL  \(\uparrow\) 
 & SR  \(\uparrow\) & SPL  \(\uparrow\) & Usage  \(\downarrow\)& SR  \(\uparrow\) & SPL  \(\uparrow\) & Usage  \(\downarrow\)
 & SR  \(\uparrow\) & SPL  \(\uparrow\) & Usage  \(\downarrow\)& SR  \(\uparrow\) & SPL  \(\uparrow\) & Usage  \(\downarrow\) \\
\midrule
0\% & 23 & 16 & 30 & 27 & 23 & 16 & 0 & 30 & 27 & 0 & 23 & 16 & 0 & 30 & 27 & 0 \\
100\% & 100 & 100 & 100 & 100 & 70 & 54 & 9.4 & 68 & 65 & 7.8 & 55 & 34 & 11.0 & 63 & 52 & 9.4 \\
\midrule 
\rowcolor{gray!40}
\multicolumn{17}{l}{\textbf{Task-wise Intervention}  ($I(s_t \rightarrow T)$) }\\
Random 10\% & 28 & 20 & 40 & 37 & 28 & 20 & 0.9 & 38 & 35 & 0.7 & 26 & 18 & 1.2 & 35 & 31 & 1.0 \\
Random 30\% & 38 & 28 & 53 & 51 & 38 & 28 & 2.8 & 42 & 39 & 2.7 & 33 & 21 & 3.3 & 38 & 34 & 2.9 \\
Random 70\% & 56 & 41 & 76 & 76 & 56 & 41 & 6.8 & 57 & 54 & 5.4 & 51 & 32 & 7.5 & 48 & 40 & 6.6 \\
\midrule
\multicolumn{17}{l}{\textbf{PRM-Thresholded: All steps} ($I(s_{[0:-1]} \rightarrow T)$)}  \\
10\% & \textbf{60} & \textbf{53} & \textbf{43} & \textbf{40} & 45 & 33 & 4.1 & 35 & 46 & 1.5 & 38 & 26 & 4.7 & 38 & 34 & 1.0 \\
30\% & \textbf{75} & \textbf{68} & \textbf{70} & \textbf{67} & 53 & 39 & 5.3 & 50 & 31 & 3.8 & 43 & 29 & 6.6 & 48 & 42 & 4.2 \\
\midrule
\multicolumn{17}{l}{\textbf{PRM-Thresholded: First five steps} ($I(s_{[0:5} \rightarrow T)$) } \\
10\% & NA & NA & NA & NA & 28 & 19 & 1.4 & 38 & 32 & 0.6 & 30 & 20 & 1.9 & 35 & 29 & 1.0 \\
30\% & NA & NA & NA & NA & 50 & 34 & 4.5 & 38 & 32 & 3.5 & 38 & 21 & 4.9 & 55 & 40 & 3.2 \\
\midrule[1pt]

\rowcolor{gray!40}
\multicolumn{17}{l}{\textbf{State-wise Intervention } ($I(s_t)$) } \\
Random 10\% & NA & NA & NA & NA & \textbf{48} & \textbf{33} & \textbf{1.2} & \textbf{48} & \textbf{38} & \textbf{1.1} & \textbf{25} & \textbf{16} & \textbf{1.0} & \textbf{43} & \textbf{37} & \textbf{1.4} \\
Random 30\% & NA & NA & NA & NA & \textbf{65} & \textbf{47} & \textbf{4.0} & \textbf{50} & \textbf{44} & \textbf{2.9} & 32.5 & 21 & 3.9 & \textbf{47} & \textbf{38} & \textbf{3.7} \\
\midrule
\multicolumn{17}{l}{\textbf{PRM-Thresholded}}  \\
10\% & NA & NA & NA & NA & 30 & 18 & 0.9 & 43 & 35 & 1.0 & 23 & 11 & 1.2 & 40 & 31 & 1.1 \\
30\% & NA & NA & NA & NA & 45 & 26 & 2.5 & 43 & 36 & 1.8 & \textbf{50} & \textbf{30} & \textbf{2.5} & 38 & 30 & 2.5 \\
\bottomrule
\end{tabular}%
}
\label{tab:big_table}
\end{table*}

\section{Detailed Derivation of Usage/Policy Iteration}
\label{app:derivation}

\paragraph{Part I: Decomposing Value Function into Success and Usage}
If \(\pi\) is any policy, then the value under \(\pi\) is the expected sum:
\[
\resizebox{1.0\columnwidth}{!}{$
V_s^\pi(r) 
\;=\;
\mathbb{E}_\pi \Biggl[
\sum_{t=0}^{\infty} \gamma^t \bigl(1_{\text{success at time }t} \;-\; r\,1_{\text{help at time }t}\bigr)
\;\Big|\; s_0 = s
\Biggr].
$}
\]
Rewriting,
\begin{align}
V_s^\pi(r) 
&= \underbrace{\mathbb{E}_\pi \!\Bigl[\sum_{t=0}^{\infty} \gamma^t \,1_{\{\text{success at }t\}}\Bigr]}_{\text{(discounted successes)}} 
\nonumber \\
&\quad -\,r\,\underbrace{\mathbb{E}_\pi \!\Bigl[\sum_{t=0}^{\infty} \gamma^t \,1_{\{\text{help at }t\}}\Bigr]}_{\text{(discounted helps)}}.
\end{align}
Hence,
\[
V_s^\pi(r)
\;=\;
S_s^\pi
\;-\;
r\,M_s^\pi(r).
\]

where we denote 

\[
S_s^\pi 
\;=\;
\mathbb{E}_\pi\Bigl[\sum_{t=0}^{\infty} \gamma^t\,1_{\text{success at time }t}\Bigr],
\quad
M_s^\pi(r)
\;=\;
\mathbb{E}_\pi\Bigl[\sum_{t=0}^{\infty} \gamma^t\,1_{\text{help at time }t}\Bigr].
\]

When taking the \(\max\) over all policies \(\pi\), we get \[V_s(r) = S_s^* - r\,M_s(r)\], where \(S_s^*\) and \(M_s(r)\) come from the \emph{optimal} policy. 

Thus, we can decompose the value function into 
\[
\underbrace{(\text{expected discounted success})}_{S_s}
\;-\;
r\,\underbrace{(\text{expected discounted helps})}_{M_s(r)},
\]

\noindent

\paragraph{Part II: Arriving at Piecewise Definition of Usage}

Now, let's substitute \(\boldsymbol{V_s(r) = S_s - r\,M_s(r)}\) into the Bellman Equation of Section :

\begin{equation}
V_s(r) 
=\;
\begin{cases}
\gamma \sum_{s'} P_{\mathrm{nohelp}}(s' \mid s)\,V_{s'}(r),
& \text{if nohelp at $s$},\\[6pt]
-r \;+\;\gamma \sum_{s'} P_{\mathrm{help}}(s' \mid s)\,V_{s'}(r),
& \text{if help at $s$},
\end{cases}
\label{eq:val}
\end{equation}

and 

\[
V_{s_{\mathrm{term}}}(r) =
\begin{cases}
1, & \text{if task success},\\[6pt]
0, & \text{if task failure}.
\end{cases}
\]

for terminal states.

As we plug in \(\boldsymbol{V_s(r) = S_s - r\,M_s(r)}\) to each case 

\textbf{1. if nohelp at  $s$}

\[
S_s - r\,M_s(r)
\;=\;
\gamma \sum_{s'}P_{\mathrm{nohelp}}(s'\mid s)\,\Bigl(S_{s'} - r\,M_{s'}(r)\Bigr).
\]

From the piecewise definition of $S_s$ (eq.~\ref{eq:S_pi_definition}), we have \(
S_s \;=\; \gamma \sum_{s'} P_{\mathrm{nohelp}}(s' \mid s)\,S_{s'},
\). Thus, 
\[
-r\,M_s(r)
\;=\;
-\,r\,\gamma \sum_{s'} P_{\mathrm{nohelp}}(s'\mid s)\,M_{s'}(r).
\]
Dividing through by \(-r\) (assuming \(r>0\)) gives
\[
M_s(r)
\;=\;
\gamma \sum_{s'} P_{\mathrm{nohelp}}(s'\mid s)\,M_{s'}(r),
\]
for the nohelp branch.

\textbf{2. The help branch.}\\
\[
S_s - r\,M_s(r)
\;=\;
-r \;+\;
\gamma \sum_{s'}P_{\mathrm{help}}(s'\mid s)\,\Bigl(S_{s'} - r\,M_{s'}(r)\Bigr).
\]
Again using the piecewise defintiion of eq.~\ref{eq:S_pi_definition} that $\gamma \displaystyle \sum_{s'} P_{\mathrm{help}}(s' \mid s)\;S_{s'}^\pi$, we can isolate the usage terms:
\[
-r\,M_s(r)
\;=\;
-\,r 
\;-\; 
r\,\gamma \sum_{s'} P_{\mathrm{help}}(s'\mid s)\,M_{s'}(r),
\]

\[ \iff\ 
r\,\Bigl(1 - \gamma \sum_{s'} P_{\mathrm{help}}(s'\mid s)\,M_{s'}(r)\Bigr)
\;=\;
r\,M_s(r).
\]
Dividing through by \(r\) and rearranging yields
\[
M_s(r)
\;=\;
1 
\;+\;
\gamma \sum_{s'} P_{\mathrm{help}}(s'\mid s)\,M_{s'}(r).
\]
Thus, in the help branch, we add \(\,1\) for the immediate usage plus the (discounted) future usages under help transitions.

Thus, we get 

\begingroup
\setlength{\fboxsep}{1pt}
\begin{equation*}
\refstepcounter{equation}
\boxed{%
M_s(r) =
\begin{cases}
\gamma \displaystyle\sum_{s'} P_{\mathrm{nohelp}}(s' \mid s)\,M_{s'}(r) \text{ if nohelp},\\[8pt]
1 \;+\; \gamma \displaystyle\sum_{s'} P_{\mathrm{help}}(s' \mid s)\,M_{s'}(r) \text{ if help}.
\end{cases}
}(\theequation)\label{eq:piecewiseM}
\end{equation*}
\endgroup

\paragraph{Part III: Arriving at Optimal Policy and Usage}

First, let's derive the threshold condition. 
From the value function Bellman Equation (eq.~\ref{eq:val}), \emph{help} is chosen iff:
\[
-r \;+\;\gamma \!\!\sum_{s'} P_{\mathrm{help}}(s'\mid s)\,V_{s'}(r)
>
\gamma \!\!\sum_{s'} P_{\mathrm{nohelp}}(s'\mid s)\,V_{s'}(r)
\]
Rewriting \(V_s(r) = S_s - r\,M_s(r)\) and isolating the cost component \(-r\) yields the \emph{threshold} condition:
\[
r 
\;<\; 
\frac{\Delta p_s}{\Delta M_s},
\]
where we denote

\[
\begin{aligned}
\Delta p_s &\;=\; p_{\mathrm{help}}(s) \;-\; p_{\mathrm{nohelp}}(s),
\\[6pt]
\Delta M_s &\;=\;
\underbrace{p_{\mathrm{help}}(s)\Bigl[1 + \gamma \sum_{s'} P_{\mathrm{help}}(s'\mid s)\,M_{s'}(r)}_{\text{help\_val}_s}\Bigr]
\\
&\quad-\;
\underbrace{p_{\mathrm{nohelp}}(s)\Bigl[\gamma \sum_{s'} P_{\mathrm{nohelp}}(s'\mid s)\,M_{s'}(r)}_{\text{nohelp\_val}_s}\Bigr].
\end{aligned}
\]

Intuitively, \(\Delta p_s\) and \(\Delta M_s\) capture how much additional \emph{success probability} vs.\ \emph{usage} we get by choosing help over nohelp at \(s\). 

Hence, we arrive at 
\begin{equation}
\boxed{
\pi_s(r) 
\;=\;
\begin{cases}
\text{\texttt{help}},   & \text{if } r < \frac{\Delta p_s}{\Delta M_s},\\[4pt]
\text{\texttt{nohelp}}, & \text{otherwise}.
\end{cases}
}
\label{eq:choose_help_condition}
\end{equation}

Now, combining eq.~\ref{eq:piecewiseM} and eq.~\ref{eq:choose_help_condition}, we get 

\begingroup
\centering
\setlength{\abovedisplayskip}{5pt}
\setlength{\abovedisplayshortskip}{0pt}
\setlength{\belowdisplayskip}{0pt}
\thinmuskip=1mu
\medmuskip=1mu
\thickmuskip=1mu
\relax
\vspace{-2em}
\[
\resizebox{1.0\columnwidth}{!}{%
\boxed{%
\begin{aligned}
M_s(r)=&
  \begin{cases}
    M_s^{\mathrm{help}}=1+\gamma\sum_{s'}P_{\mathrm{help}}(s'\mid s)M_{s'}(r),
      & \text{if }\pi(s)=\text{help},\\[4pt]
    M_s^{\mathrm{nohelp}}=\gamma\sum_{s'}P_{\mathrm{nohelp}}(s'\mid s)M_{s'}(r),
      & \text{otherwise}.
  \end{cases}\\[4pt]
\pi(s)=&\text{help} \;\iff\; r<\frac{\Delta p_s}{\Delta M_s}.
\end{aligned}%
}%
}
\]
\endgroup

\noindent 
\paragraph{Lemma: Piecewise Definition of \(\boldsymbol{S_s}\) (Help vs.\ Nohelp).}
If \(S_s\) is interpreted as the \emph{discounted probability of eventually reaching success} under a policy \(\pi\) that may choose either \(\mathrm{help}\) or \(\mathrm{nohelp}\), we can write
\[
S_s^\pi 
\;=\;
\mathbb{E}_\pi \Bigl[\,
\sum_{t=0}^{\infty} \gamma^t\,1_{\{\text{state at time }t \text{ is success}\}}
\;\Big|\; s_0=s
\Bigr].
\]
In an MDP setting with two possible actions, \(\mathrm{help}\) or \(\mathrm{nohelp}\), the policy \(\pi\) dictates which action to take at each state \(s\).  Correspondingly, the recursion for \(S_s^\pi\) becomes:
\begin{equation}
\label{eq:S_pi_definition}
\resizebox{1.0\columnwidth}{!}{%
$S_s^\pi
\;=\;
\begin{cases}
1, 
& \text{if \(s\) is a terminal success state},\\[6pt]
0, 
& \text{if \(s\) is a terminal failure state},\\[6pt]
\gamma \displaystyle \sum_{s'} P_{\mathrm{nohelp}}(s' \mid s)\;S_{s'}^\pi,
& \text{if \(\pi\) chooses \texttt{nohelp} at \(s\)},\\[6pt]
\gamma \displaystyle \sum_{s'} P_{\mathrm{help}}(s' \mid s)\;S_{s'}^\pi,
& \text{if \(\pi\) chooses \texttt{help} at \(s\)}.
\end{cases}$
}
\end{equation}
\begin{itemize}
\item \textbf{If \(s\) is success:} We set \(S_s^\pi = 1\). This means that if you start in a success state, the probability of ``having achieved success'' (discounted or not) is exactly 1.

\item \textbf{If \(s\) is nonterminal:} Then there is no immediate success contribution at \(s\) itself, and we simply recurse to the next state via either \(P_{\mathrm{nohelp}}(\cdot\mid s)\) or \(P_{\mathrm{help}}(\cdot\mid s)\), multiplied by the factor \(\gamma\). Thus, no explicit \(\mathbf{1}\{\text{$s$ is success}\}\) is needed inside the sum, because we have already distinguished the success case in the first line of the piecewise definition.
\end{itemize}

Thus, under a given policy \(\pi\), each nonterminal state \(s\) follows whichever transition probabilities (\(\mathrm{nohelp}\) or \(\mathrm{help}\)) \(\,\pi\) prescribes at that state. The boundary condition \(S_{s_{\mathrm{term}}}^\pi = 1\) applies to all terminal success states.

\section{Proof of Convergence}
\label{app:convergence}
In Appendix~\ref{app:derivation}, we showed that the boxed equation 

\begingroup
\centering
\setlength{\abovedisplayskip}{5pt}
\setlength{\abovedisplayshortskip}{0pt}
\setlength{\belowdisplayskip}{0pt}
\thinmuskip=1mu
\medmuskip=1mu
\thickmuskip=1mu
\relax
\vspace{-2em}
\[
\resizebox{1.0\columnwidth}{!}{%
\boxed{%
\begin{aligned}
M_s(r)=&
  \begin{cases}
    M_s^{\mathrm{help}}=1+\gamma\sum_{s'}P_{\mathrm{help}}(s'\mid s)M_{s'}(r),
      & \text{if }\pi(s)=\text{help},\\[4pt]
    M_s^{\mathrm{nohelp}}=\gamma\sum_{s'}P_{\mathrm{nohelp}}(s'\mid s)M_{s'}(r),
      & \text{otherwise}.
  \end{cases}\\[4pt]
\pi(s)=&\text{help} \;\iff\; r<\frac{\Delta p_s}{\Delta M_s}.
\end{aligned}
}
}
\]
\endgroup

is equivalent to the standard Bellman recursion for
\[
V^{\pi}_s(r) 
=\;
\begin{cases}
\gamma \sum_{s'} P_{\mathrm{nohelp}}(s' \mid s)\,V^{\pi}_{s'}(r), & \text{if nohelp}, \\[4pt]
-r \;+\;\gamma \sum_{s'} P_{\mathrm{help}}(s' \mid s)\,V^{\pi}_{s'}(r), & \text{if help}.
\end{cases}
\]
where
\[
V^{\pi}_{s_{\mathrm{term}}}(r) =
\begin{cases}
1, & \text{if task success},\\[4pt]
0, & \text{if task failure}.
\end{cases}
\]

Because solving value iteration for the above $V^{\pi}_s(r)$ converges to a unique fixed point $V_s(r)$ and the corresponding policy $\pi^*(r)$, we know that the iteration of the boxed equation also converges to a unique fixed point \(M^*(r)\) and $\pi^*(r)$.

\section{Details on Extensions to Multiple Interventions}
\label{app:multiple}

Our algorithm naturally extends to multiple intervention types, as explained in Sec.~\ref{sec:method_multi}. We explain the details.

\subsection{Formulation and Reward Regime}

We consider a stochastic process with states \(s \in \mathcal{S}\) and transition probabilities \(P(s' \mid s)\). At any \emph{non-terminal} state \(s\), the agent may choose from multiple interventions 
\(\{\text{help1}, \text{help2}, \dots, \text{help}K\}\) or \(\text{nohelp}\). 
Each intervention \(\text{help}_i\) can improve the probability of success at the cost of incurring usage.  Conversely, \(\text{nohelp}\) avoids usage costs but may have a lower chance of success. We aim to \emph{maximize} the task success rate while keeping the expected discounted number of each intervention (or total usage) below a certain budget \(C\). 

Concretely, let \(\gamma \in (0,1)\) be the discount factor. Suppose from an initial state \(s_0\), we want
\[
\begin{aligned}
& \mathbb{E}\bigl[\text{(Success)}\bigr]
\\
\text{subject to}\quad
& \mathbb{E}\Bigl[\sum_{t=0}^{\infty} \gamma^t\,\#(\text{helps at time }t)\Bigr]
\;\le\; C.
\end{aligned}
\]
One can equivalently encode this via a \textit{cost} \((r_1, r_2, \dots, r_K)\) for each intervention \(\text{help}i\), or treat it via a usage-based dynamic programming approach. Below, we use a \textbf{reward regime} that translates each help call into a negative reward. This allows standard value-iteration (VI) or usage-based iteration for an MDP with multiple interventions.

\paragraph{Reward Regime.}
At each non-terminal state \(s\), the agent chooses among:
\[
\text{Actions} \;=\; \{\text{nohelp}, \,\text{help}_1, \,\dots, \text{help}_K\}.
\]
The immediate reward is:
\begin{itemize}[leftmargin=*]
    \item \(\text{help}_i\): a reward of \(-r_i\).
    \item \(\text{nohelp}\): a reward of \(0\).
    \item \textit{Terminal States:} success yields a reward of \(+1\), failure yields \(0\).
\end{itemize}
Hence, if an agent eventually succeeds, it gains \(+1\) minus the sum of costs \(\sum_i r_i\) times the discounted number of times each \(\text{help}_i\) was used. 

\paragraph{Notation for Success Probabilities.}
When analyzing usage or success, we often use a \emph{probability-of-success} model:
\[
p_{\text{nohelp}}(s) \;=\; \Pr(\text{success}\mid s, \text{nohelp}), \]
\[p_{\text{help}_i}(s) \;=\; \Pr(\text{success}\mid s, \text{help}_i).
\]
We likewise denote state transition kernels 
\(P_{\text{nohelp}}(s'|s)\) or \(P_{\text{help}_i}(s'|s)\)
to capture the distribution over next states under each chosen action.

\subsection{Derivation of Usage/Policy Iteration for Multiple Interventions}

\paragraph{Overview.}
We start from value iteration, as in Sec.~\ref{sec:method_derivation}. The value function is:
\[
\begin{aligned}
V(s) 
&= \max\Bigl\{
  0 + \gamma \sum_{s'} P_{\mathrm{nohelp}}(s'|s)\,V(s'),\\
&\qquad \{-r_i + \gamma \sum_{s'} P_{\mathrm{help}i}(s'|s)\,V(s')\}_{i=1}^K
\Bigr\}.
\end{aligned}
\]

with \(V(s_{\mathrm{success}})=1\) and \(V(s_{\mathrm{failure}})=0\) for terminal states. Now, we drive a \textit{usage-based} DP from this:
\[
\begin{aligned}
M_s^i 
&= \text{expected discounted \# of times we use intervention } i,\\
&\quad \text{starting from } s.
\end{aligned}
\]

If we pick \(\text{help}_i\) in state \(s\), then
\[
M_s^i(\text{help}_i) = 1 + \gamma \!\sum_{s'} P_{\mathrm{help}_i}(s'|s)\;M_{s'}^i,\]
\[M_s^j(\text{help}_i) = 0 + \gamma \!\sum_{s'} P_{\mathrm{help}_i}(s'|s)\;M_{s'}^j,
\]
for \(j\neq i\).  
If we pick \(\text{nohelp}\),
\[
M_s^i(\text{nohelp})
\;=\;
\gamma\sum_{s'} P_{\mathrm{nohelp}}(s'|s)\;M_{s'}^i,
\quad
\forall i=1,\dots,K.
\]

\paragraph{Threshold Conditions for Multiple Interventions.}
In the single-intervention case, we derived \(\mathrm{ratio} = \dfrac{\Delta \text{success}}{\Delta \text{usage}}\). For multiple interventions, each \(\text{help}_i\) has:
\[
\Delta p^i_s 
\;=\;
p_{\mathrm{help}_i}(s) \;-\; p_{\mathrm{nohelp}}(s),\]
\[
\Delta M^i_s 
\;=\;
p_{\mathrm{help}_i}(s)\,\bigl[M_s^i(\text{help}_i)\bigr]
\;-\;
p_{\mathrm{nohelp}}(s)\,\bigl[M_s^i(\text{nohelp})\bigr].
\]
We say \(\text{help}_i\) is cost-effective (vs.\ nohelp) if
\[
\mathrm{ratio}_i(s)
\;=\;
\frac{\Delta p^i_s}{\Delta M^i_s}
\;>\;r_i.
\]
If \(\mathrm{ratio}_i(s)\le r_i\) for all \(i\), we choose \(\text{nohelp}\).  If exactly one \(\text{help}_i\) is cost-effective, we pick \(\text{help}_i\). If \emph{multiple} helps pass the ratio test, we pick whichever yields the smallest \textit{combined} cost
\[
\bigl(r_1\,M_s^1,\,r_2\,M_s^2,\,\dots,r_K\,M_s^K\bigr)
\Longrightarrow
\text{minimize }\,\sum_{i=1}^{K} r_i\,M_s^i(\text{help}_i).
\]
These local decisions define a \emph{policy update} at each state \(s\).  Iterating the usage functions \(\{M_s^i\}\) and reselecting among \(\{\text{help}_i, \text{nohelp}\}\) converges to a stable fixed point.  This final stable policy is \(\pi^*\).  This \(\pi^*\) is exactly the same solution a standard value iteration approach (with reward \(\{-r_i\}\)) would find, with arguments similar to Appendix~\ref{app:convergence}.

\subsection{Algorithm}

The algorithm for the multiple intervention setting are in three main phases:

\textbf{Phase 1: Data Collection and Transition Model.}
  \begin{itemize}[leftmargin=1em]
  \item Collect transitions offline by running partial-rollouts with \(\{\text{help}i, \text{nohelp}\}\) chosen randomly or by partial heuristics.
  \item Maintain counts \(\texttt{count}[s][a][s']\) for each action \(a\in\{\text{help}1,\dots,\text{help}K,\text{nohelp}\}\).
  \item Estimate \[\hat{P}(s'|s,a)=\texttt{count}[s][a][s']/\sum_{x}\texttt{count}[s][a][x],\] for $a \in \{\text{nohelp}, \,\text{help}_1, \,\dots, \text{help}_K\}$
  \end{itemize}

\textbf{Phase 2: Offline Usage/Policy Iteration (Multiple Interventions).}
First, initialize usage counters \(\{M_s^i\}_{i=1}^K\) to zero (or any guess).
Then, Repeat until convergence:
    \begin{enumerate}[leftmargin=1.6em]
      \item \emph{Compute usage for each action}:
        \[
        \begin{aligned}
          M_s^i(\text{help}j) &=
            \begin{cases}
              1 + \gamma \sum_{s'}P_{\mathrm{help}j}(s'|s)\,M_{s'}^i,
                & \text{if }i=j,\\
              \gamma \sum_{s'}P_{\mathrm{help}j}(s'|s)\,M_{s'}^i,
                & \text{if }i\neq j,
            \end{cases}\\
          M_s^i(\text{nohelp}) &=
            \gamma \sum_{s'}P_{\mathrm{nohelp}}(s'|s)\,M_{s'}^i.
        \end{aligned}
        \]
      \item \emph{Compute $\Delta p^i_s$ and $\Delta M^i_s$}, then check the ratio test $\mathrm{ratio}_i(s)>\!r_i$. 
      \item \emph{Policy update}: 
        \[
        \begin{aligned}
        \pi_s 
        &= \arg\min_{a\in\{\text{help}1,\dots,\text{help}K,\text{nohelp}\}}
        \Bigl\{
          \sum_{i=1}^K r_i\,M_s^i(a)
        \Bigr\}\\
        &\quad \text{subject to}\quad \mathrm{ratio}_i > r_i.
        \end{aligned}
        \]
      \item \emph{Update counters}: 
        \(\displaystyle M_s^i\;\leftarrow\;M_s^i(\pi_s).\)
      \item \emph{Check convergence}: if \(\max_{s,i}\bigl|M_s^i-\text{old}_s^i\bigr| < \varepsilon\), stop.
    \end{enumerate}
 
Finally, we output the stable usage counters \(\{M_s^i\}_{i=1}^K\) and the final policy \(\pi^*\).

\textbf{Phase 3: Final Policy Representation (SFT or Other).}
  \begin{itemize}[leftmargin=1em]
  \item We store the final help/nohelp decisions in a table \(\pi^*(s)\). 
  \item For states $s$ in the training data, we know exactly which action the usage-based DP prescribes.
  \item Train the actual “helper” model (e.g. a neural policy or large language model) via \emph{supervised finetuning} to mimic $\pi^*(s)$ on the collected states $s$.
  \end{itemize}

\paragraph{Relation to Single-Intervention Case.}
If $K=1$, the above steps reduce exactly to the single-intervention usage-based iteration.  
If $r_1=r_2=\cdots=r_K=r$, then each help has the same cost, and we can unify them if needed.

\end{document}